\documentclass{article}

\usepackage{microtype}
\usepackage{graphicx}
\usepackage{subcaption}
\usepackage{booktabs} 

\usepackage{hyperref}

\usepackage[preprint]{icml2026}

\usepackage{amsmath}
\usepackage{amssymb}
\usepackage{mathtools}
\usepackage{amsthm}

\usepackage[capitalize,noabbrev]{cleveref}

\theoremstyle{plain}

\theoremstyle{definition}

\theoremstyle{remark}

\usepackage[textsize=tiny]{todonotes}

\usepackage[dvipsnames]{xcolor}
\usepackage{natbib}

\usepackage{graphicx}
\usepackage{subcaption}

\usepackage{amsmath}
\usepackage{amssymb}
\usepackage{mathtools}
\usepackage{amsthm}
\usepackage{bbm}
\usepackage{pifont}
\usepackage{yhmath} 

\usepackage{algorithm}
\usepackage{algorithmic}

\usepackage{enumitem}

\newcommand{\cS}{\mathcal{S}}
\newcommand{\cD}{\mathcal{D}}
\newcommand{\cN}{\mathcal{N}}

\newcommand{\cA}{\mathcal{A}}

\newcommand{\bbE}{\mathbb{E}}

\newcommand{\bbR}{\mathbb{R}}
\newcommand{\bbP}{\mathbb{P}}

\newcommand{\qe}{\text{\tiny query}}

\newcommand{\te}{\text{\tiny test}}
\newcommand{\dppt}{$\mathrm{DP^2T}$}

\DeclareMathOperator*{\KL}{\mathrm{KL}}

\DeclareMathOperator*{\argmax}{argmax}

\icmltitlerunning{Learning in Context, Guided by Choice: A Reward-Free Paradigm for Reinforcement Learning with Transformers}

\begin{document}

\twocolumn[
  \icmltitle{Learning in Context, Guided by Choice:\\ 
A Reward-Free Paradigm for Reinforcement Learning with Transformers}



  \icmlsetsymbol{equal}{*}

\begin{icmlauthorlist}
\icmlauthor{Juncheng Dong}{ee}
\icmlauthor{Bowen He}{cs}
\icmlauthor{Moyang Guo}{ee}
\icmlauthor{Ethan X. Fang}{ef}
\icmlauthor{Zhuoran Yang}{zy}
\icmlauthor{Vahid Tarokh}{ee}
\end{icmlauthorlist}

\icmlaffiliation{ee}{Department of Electrical and Computer Engineering, Duke University, Durham, US}
\icmlaffiliation{cs}{Department of Computer Science, Duke University, Durham, US}
\icmlaffiliation{ef}{Department of Biostatistics and Bioinformatics, Duke University, Durham, US}
\icmlaffiliation{zy}{Department of Statistics and Data Science, Yale University,
New Haven, US}
\icmlcorrespondingauthor{Juncheng Dong}{juncheng.dong@duke.edu}

  \icmlkeywords{Machine Learning, ICML}

  \vskip 0.3in
]



\printAffiliationsAndNotice{}  

\begin{abstract}
In-context reinforcement learning (ICRL) leverages the in-context learning capabilities of transformer models (TMs) to efficiently generalize to unseen sequential decision-making tasks without parameter updates. 
However, existing ICRL methods rely on explicit reward signals during pretraining, which limits their applicability when rewards are ambiguous, hard to specify, or costly to obtain.
To overcome this limitation, we propose a new learning paradigm, \emph{In-Context Preference-based Reinforcement Learning} (ICPRL), in which both pretraining and deployment rely solely on preference feedback, eliminating the need for reward supervision.
We study two variants that differ in the granularity of feedback: \emph{Immediate Preference-based RL} (I-PRL) with per-step preferences, and \emph{Trajectory Preference-based RL} (T-PRL) with trajectory-level comparisons. 
We first show that supervised pretraining, a standard approach in ICRL, remains effective under preference-only context datasets, demonstrating the feasibility of in-context reinforcement learning using only preference signals. 
To further improve data efficiency, we introduce alternative preference-native frameworks for I-PRL and T-PRL that directly optimize TM policies from preference data without requiring reward signals nor optimal action labels.
Experiments on dueling bandits, navigation, and continuous control tasks demonstrate that ICPRL enables strong in-context generalization to unseen tasks, achieving performance comparable to ICRL methods trained with full reward supervision.
\end{abstract}

\section{Introduction}

\begin{figure*}
  \centering
  \includegraphics[width=0.8\textwidth]{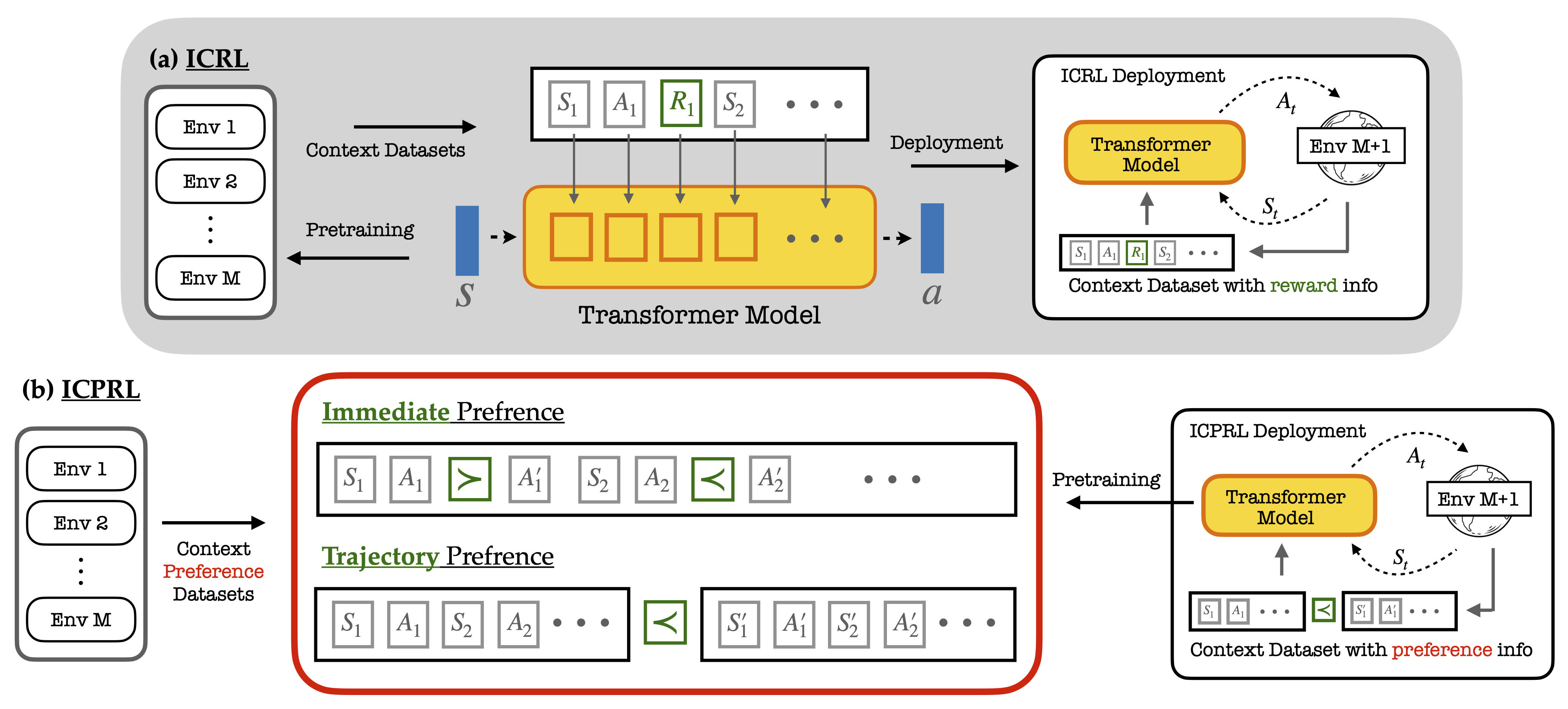}
  \caption{\textbf{(a)} ICRL methods adapt to \emph{\color{Black}new} RL tasks using in-context learning, but both their pretraining and deployment require access to reward signals, which can be costly or impractical to obtain in many settings. \textbf{(b)} This work proposes a novel ICRL paradigm that uses \emph{\color{Black}only preference data} for both pretraining and deployment, eliminating the need for explicit reward supervision.}
  \label{fig:intro}
\end{figure*}

Reinforcement learning (RL) has achieved impressive successes across a wide range of domains, including robotics~\citep{kober2013reinforcement}, recommendation systems~\citep{afsar2022reinforcement}, and more recently, post-training of large language models~\citep{ziegler2019fine}.  
Despite these advances, both online and offline RL methods remain highly data-intensive, typically requiring a large number of environment interactions to achieve satisfactory performance.
To address this issue, \emph{in-context reinforcement learning} (ICRL) has recently gained attention~\citep{AD}. Building on the in-context learning capabilities of transformer models ({TMs}), ICRL methods pretrain TMs on a diverse set of RL tasks and deploy them directly to \emph{new} tasks at test time. See Figure~\ref{fig:intro} for visuals. 
These pretrained TMs act as meta-policies, rapidly adapting to unseen tasks using only a small number of trajectories and \emph{without} parameter updates. 
As a result, ICRL achieves strong generalization with significantly fewer interactions than conventional RL learners~\citep{DPT,dong2025incontext}.

\textbf{Challenge: Reward Dependence in ICRL.} 
Despite their promise, existing ICRL methods critically rely on access to \emph{explicit reward signals} during both pretraining and deployment. 
This reliance limits applicability in many real-world settings where rewards are ambiguous, difficult to specify, or expensive to obtain~\citep{ibarz2018reward}. 
Moreover, ICRL exacerbates the reward design problem: to enable generalization across tasks, reward signals must be not only informative but also \emph{semantically consistent across environments}, a stringent assumption  in practice.

\textbf{Preference Feedback as An Alternative.}
An increasingly common alternative to explicit reward supervision is \emph{preference feedback}, where learning is guided by relative comparisons rather than scalar rewards~\citep{christiano2017deep}. Preferences are often easier to elicit, more robust to reward misspecification, and naturally invariant to task-specific reward scaling. 
Importantly, preference feedback provides a form of supervision that is well aligned with the objectives of ICRL: it conveys task-relevant information without requiring a shared reward function across tasks.

\textbf{A Reward-Free Paradigm for ICRL.} In this light, we propose \emph{In-Context Preference-based Reinforcement Learning} (\textbf{ICPRL}), a new paradigm for ICRL in which the learning algorithm operates \emph{exclusively} on preference feedback during both pretraining and deployment. Throughout, \emph{reward-free} refers to the supervision available to the learner: the model never observes scalar rewards. As is standard in preference-based RL, synthetic preferences may be generated from a hidden reward function for controlled evaluation, but this reward is never accessed by the learning algorithm. Within the ICPRL paradigm, we study two settings that differ in the granularity of preference feedback: \emph{Immediate Preference-based RL} (\textbf{I-PRL}), which learns from per-step preferences, and \emph{Trajectory Preference-based RL} (\textbf{T-PRL}), which relies on comparisons between complete trajectories.

\textbf{Contributions.} Our main contributions are as follows.
\begin{itemize}[left=0.5em]
    \item We introduce ICPRL, a preference-only paradigm for in-context reinforcement learning that removes the need for explicit reward supervision while retaining rapid in-context adaptation and generalization.
    \item We instantiate ICPRL in both immediate and trajectory-level preference settings and propose a simple supervised framework that trains a TM to predict optimal actions for query states, conditioned on context datasets constructed from preference data. We show that \emph{supervised} in-context pretraining, a well-established framework for ICRL~\citep{DPT}, remains effective even under preference-only feedback. 
    \item To eliminate reliance on optimal action labels, we further propose \emph{preference-native} training frameworks for both I-PRL and T-PRL that directly exploit preference structure during pretraining.
\end{itemize}

\textbf{Empirical evaluation.}
We evaluate ICPRL across a diverse suite of tasks, including dueling bandits~\citep{yue2012k}, navigation~\citep{AD}, and continuous control~\citep{yu2020meta}. Despite operating exclusively on preference feedback, our models generalize effectively to unseen tasks and achieve competitive performance relative to reward-supervised ICRL baselines.

\section{Related Work}\label{sec:related-work}
We discuss the most relevant related work, deferring the full literature review to Appendix~\ref{sec:app-related-work} .

\textbf{Preference-based Reinforcement Learning (PbRL).}
PbRL focuses on learning from preference signals, typically given as comparisons between actions or trajectories, instead of relying on scalar reward functions~\citep{christiano2017deep, wirth2017survey, brown2019extrapolating}. 
A common paradigm in PbRL involves a two-stage model-based process: first, learning a reward function from preferences, and second, optimizing a policy using standard RL algorithms with the learned reward~\citep{ibarz2018reward,  pmlr-v139-lee21i, liu2023efficient}. 
The most salient distinction between the proposed ICPRL paradigm and prior PbRL approaches is that PbRL assumes the \emph{same} environment for training and evaluation, whereas ICPRL operates in the in-context setting, where a model must generalize to \emph{new} tasks at inference time without parameter updates.


\textbf{Transformer Models and In-Context Reinforcement Learning.} Transformer models have demonstrated superior performance in RL problem~\citep{li2023a,yuan2023transformer}. Building on the in-context learning abilities of transformers, ICRL methods aim to learn a TM-based meta-policy to generalize to unseen tasks in context. ICRL methods differ in their requirements of context datasets. For example, \emph{Algorithm Distillation} (AD)~\citep{AD} and its variations~\citep{zisman2023emergence,tarasov2025yes} uses sequential modeling to emulate the learning process of RL algorithms, i.e., meta-learning~\citep{vilalta2002perspective}. Decision transformer~\citep{chen2021decision}(DT)-based methods rely on return-to-go to guide the transformer models to generalize to new tasks~\citep{grigsby2023amago,huang2024context,schmied2024retrieval}. A recent framework \emph{Decision Pretrained Transformer} (DPT) uses supervised-pretraining for in-context decision making. DPT trains transformers to predict the optimal action given a query state and a set of transitions. Despite its strong performance, DPT requires high-quality action labels for its pretraining. However, to our best knowledge, all existing ICRL methods assume explicit reward (goal) signal for pretraining. To this end, we make the first step towards reward/goal-free ICRL.

\section{Preliminary}

\textbf{Markov Decision Process (MDPs).} We consider standard episodic RL problem, defined as MDPs. 
An MDP $\tau$ is a tuple $(\cS, \cA, P_{\tau}, R_{\tau}, \rho_{\tau}, H)$ where $\cS$ and $\cA$ are the sets of all possible states and actions, $P_\tau:\cS\times\cA \rightarrow \Delta(\cS)$ is the distribution of the next state, $R_{\tau}:\cS\times\cA \rightarrow \bbR$ is the reward function, $\rho_\tau \in \Delta(\cS)$ is the initial state distribution, and $H$ is the episode horizon. At the initial step $h=1$, an initial state $s_1 \in \cS$ is sampled according to $\rho_\tau$. At each time step $h$, the agent chooses action $a_h \in \cA$ and receives reward $r_h = R_{\tau}(s_h,a_h)$. Then the next state $s_{h+1}$ is generated following $P_\tau(s_h,a_h)$. A policy $\pi: \cS \rightarrow \Delta(\cA)$ maps the current state to an action distribution. Let $G_{\tau}(\pi) = \bbE[\sum_{h=1}^{H}r_h|\pi,\tau]$ denote the expected cumulative reward of $\pi$ for task $\tau$. The goal of an agent is to learn the optimal policy $\pi_\tau^{\star}$ that maximizes $G_{\tau}(\pi)$. We highlight that, throughout this paper, rewards are only used to define the underlying task and for evaluation purposes; in ICPRL the learning algorithm does not observe $R_\tau$ and instead receives only preference feedback.

\textbf{ICRL.} ICRL methods first pretrain a transformer $T_\theta$ on diverse RL tasks $\{\tau_i\}^m_{i=1}$, where each $\tau_i \sim p_\tau$ is an MDP drawn independently from the task distribution $p_\tau$. 
The pretrained policy $T_\theta$ is then deployed on a new task $\tau^\te \sim p_\tau$\footnote{We assume the test task distribution matches the training one.}.
The {\bf goal} is to pretrain $T_\theta$ so it generalizes \emph{in context} to $\tau^\te$ during deployment. 
To facilitate in context generalization, existing ICRL methods typically provide a context dataset $D^R$ with information about $\tau^\te$, and the policy $T_\theta(a |s, D^R)$ outputs actions $a$ for given states $s$, conditioned on $D^R$. 
In other words, $T_\theta$ acts as a meta-policy, adapting to tasks conditioned on the context datasets collected from these tasks.
Different ICRL methods assume different dataset formats. For example, DPT assumes $D^R = \{s_j, a_j , r_j, s'_j\}_{j \in [H]}$, a set of transitions collected by a behavior policy~\citep{DPT}; policy distillation methods like AD assume $D^R=\{\xi_{i,j}\}^J_{j=1}$, a set of trajectories $\xi_{i,j}$ collected by increasingly improving policies~\citep{AD}. Despite these format differences, existing ICRL methods typically assume that context datasets include explicit reward signals. In this work, we preserve the in-context adaptation structure of ICRL while replacing reward-based context with preference-based context.

\textbf{Supervised Pretraining for ICRL.} One representative method for ICRL is DPT, which is the pioneering work that proposes to use supervised pretraining for ICRL. DPT assumes a pretraining dataset $\cD_i = \{s^\qe_i, a_i^{\star}, D^R_i\}$ for each pretraining task $\tau_i$ where $D^R_i$ is the context dataset introduced above, $s_i^\qe$ is a query state to be used for supervised pretraining, and $a_i^\star \sim \pi_{\tau_i}^\star(s_i^\qe)$ is an associated optimal action label for $s_i^\qe$ sampled from the optimal policy $\pi_{\tau_i}^\star$ for $\tau_i$~\citep{DPT}. DPT pretrains the TM policy $T_\theta$ to predict the optimal actions for query states given the context datasets with the following classification-like objective:
\begin{equation}\label{eqn:dpt-obj}
    \max_{\theta} \frac{1}{m}\sum^m_{i=1} \log T_\theta(a_i^\star|s_i^\qe, D^R_i).
\end{equation}
To better understand why DPT is effective for ICRL and to help motivate our proposed frameworks, we provide detailed insights into the mechanisms and intuitions behind supervised pretraining. Due to space constraints, see Appendix~\ref{sec:app-intuition}. While effective, this formulation relies on access to optimal action labels and reward-based context, motivating our preference-only alternatives introduced in Section~\ref{sec:algo_without_optimal_labels}.


\section{In-Context Preference-based Reinforcement Learning}
\label{sec:icprl}

\textbf{Notations.}
Let $\sigma(x) = 1/(1+\exp(-x))$ denote the sigmoid function.
We use $\succ$ to represent preference relations.
For actions, $a \succ a'$ indicates that $a$ is preferred over $a'$ in a given state;
for trajectories, $\xi \succ \xi'$ indicates that trajectory $\xi$ is preferred to $\xi'$.
We note that preferences are treated as primitive supervision signals and may be generated by humans or synthetic procedures depending on the setting.

\subsection{Problem Setup}

\textbf{Overview.}
We study two preference-based in-context learning paradigms:
\textbf{(i)} I-PRL, which receives step-wise preferences,
and \textbf{(ii)} T-PRL, with trajectory-level comparisons.
In both I-PRL and T-PRL, the agent does not observe scalar rewards.
Instead, it receives preference feedback in the form of pairwise comparisons.
The objective of ICPRL is to
pretrain a transformer-based policy that can adapt to a new task using only a small number of preference observations provided at deployment.

The key distinction between I-PRL and T-PRL lies in the granularity of preference
feedback and the resulting structure of the context dataset provided to the model.
We denote the context datasets by $D^I$ for I-PRL and $D^T$ for T-PRL.

\textbf{Latent Task Objective.}
For each task $\tau$, we assume the existence of an underlying task objective that determines the preferred behaviors.
This objective is not observed by the agent and need not be explicitly represented as a reward function.
For evaluation purposes only, we may associate $\tau$ with a latent reward function $R_\tau$, which is never revealed to the learning algorithm.
All learning and adaptation in ICPRL is driven exclusively by preference feedback.

\textbf{I-PRL Context.}
In I-PRL, preference feedback is provided at the level of individual decision points.
For a task $\tau$, the context dataset takes the form
\begin{equation}
D^I = \{ s_1,(a_1,a'_1,y_1),\dots,s_H,(a_H,a'_H,y_H) \},
\end{equation}
where $s_1 \sim \rho_\tau$, actions $a_h$ and $a'_h$ are proposed at state $s_h$,
and $y_h \in \{0,1\}$ indicates which action is preferred.
The interaction protocol specifies how the next state is selected after a comparison;
this can follow standard dueling bandit~\citep{bengs2021preference} extensions of MDPs or other task-specific mechanisms.

\textbf{T-PRL Context.}
In T-PRL, preference feedback is provided over entire trajectories.
For a task $\tau$, the context dataset is
\begin{equation}
D^T = \{ \xi, \xi', y \},
\end{equation}
where $\xi$ and $\xi'$ are two state-action trajectories of horizon $H$ collected
by possibly different behavioral policies, and $y \in \{0,1\}$ indicates which
trajectory is preferred.
Trajectories do not include reward annotations.

\textbf{Preference Modeling for Evaluation.}
To enable controlled benchmarking, we may specify a synthetic data-generation process for preference labels.
Following common practice in preference-based RL, we use the Bradley--Terry model to relate preferences to a latent reward function when generating synthetic labels~\citep{bradley1952rank}.
For I-PRL, preferences may depend on relative action quality at a state; for T-PRL, they may depend on cumulative trajectory return.
Crucially, these models describe how preference labels are generated in experiments and are not assumptions made by the learning algorithm.
ICPRL requires access only to preference comparisons and does not assume knowledge of the latent reward, optimal policy, or advantage function.

\subsection{Practicality of ICPRL}

\textbf{Reallocating the Preference Budget.}
In conventional preference-based RL, a large number of pairwise comparisons is required
to learn policies for single tasks~\citep{christiano2017deep}.
ICPRL redistributes a comparable total annotation budget across a diverse set of
pretraining tasks, enabling a pretrained policy to adapt to new tasks with only a
small number of additional comparisons at deployment.
For the same overall labeling effort, ICPRL yields policies that generalize to
previously unseen tasks, thus providing substantial practical value.

\textbf{Low-Cost Annotators.}
Recent work demonstrates that large language models (LLMs) and vision--language models
(VLMs) can act as preference annotators when prompted with task goals, enabling
inexpensive collection of trajectory-level and even step-wise preferences~\citep{klissarov2023motif,lee2023rlaif}.
While the use of LLMs as annotators is not an assumption of ICPRL, it represents a promising instantiation that can reduce human labeling cost.
We include a pilot study in Appendix~\ref{app:llm} to verify that modern LLMs can indeed identify preferred trajectories in ICRL benchmarks.

\textbf{Practical Value of I-PRL.}
Although step-wise preferences may appear costly, I-PRL is well suited to settings
where instantaneous decisions can be evaluated automatically or cheaply.
In particular, it subsumes the dueling bandit problem~\citep{yue2012k}, which has
broad applications in recommendation and online decision-making.
Step-wise feedback is information-dense, improves credit assignment, and often reduces
the number of trajectories required for effective learning.
By including both I-PRL and T-PRL, ICPRL supports a \emph{comprehensive} range of feedback regimes, from weak
trajectory comparisons to richer structured supervision.

\section{Supervised Pretraining for ICPRL}
\label{sec:proposed-frameworks}

We begin by studying a supervised pretraining approach for ICPRL that mirrors
existing ICRL methods while replacing reward-based context with preference-based
context.
Our goal in this section is not to propose a fully preference-native training
procedure, but rather to establish a diagnostic baseline that isolates the
informational content of preference-only context datasets.

\textbf{Decision Preference Pretrained Transformer (DPPT).}
We introduce \emph{Decision Preference Pretrained Transformer} (\dppt), a direct
extension of the DPT framework to the preference-only setting.
In DPT, each pretraining task $\tau_i$ is associated with a context dataset
$D_i^R$ containing reward-annotated transitions or trajectories.
In contrast, \dppt\ replaces $D_i^R$ with a preference-based context dataset
$D_i$, which may take the form of $D_i^I$ (step-wise preferences) or $D_i^T$
(trajectory-level preferences), as defined in Section~\ref{sec:icprl}.

As in DPT, \dppt\ trains a transformer policy $T_\theta(a \mid s, D)$ that maps a
query state $s$ and a context dataset $D$ to an action distribution.
Crucially, during both pretraining and deployment, the policy observes only
states and preference-based context; it never observes rewards.

\textbf{Supervised Objective.}
To probe whether preference-based context contains sufficient information for task inference and in-context adaptation, \dppt\ adopts a supervised training objective.
For each pretraining task $\tau_i$, we assume access to a query state $s_i^{\mathrm{qe}}$ and an associated target action $a_i^\star$. These target actions are used \emph{only during offline pretraining} and are
never provided to the model at deployment.
Formally, \dppt\ optimizes the same objective as DPT:
\begin{equation}
\max_{\theta} \frac{1}{m} \sum_{i=1}^m \log T_\theta(a_i^\star \mid s_i^{\mathrm{qe}}, D_i),
\end{equation}
where the distinction from DPT lies entirely in the structure of the context
dataset $D_i$, which contains preference feedback instead of reward signals.

\textbf{Interpretation.}
Unlike DPT, \dppt\ must infer task-relevant structure solely from preference
comparisons rather than explicit reward annotations.
As a result, successful generalization under \dppt\ demonstrates that
preference-only context can support in-context task inference, even when trained
using strong supervision offline.
We emphasize that \dppt\ is not fully preference-native, as it relies on target
action labels during pretraining.
Instead, it serves as a controlled baseline that motivates the development of
training frameworks that eliminate both reward signals and optimal action labels,
which we introduce in the following sections.

Despite this limitation, our experiments in Section~\ref{sec:exp} show that
\dppt\ achieves strong generalization in both I-PRL and T-PRL settings, indicating
that preference-based context provides a viable foundation for in-context
decision-making.

\section{Preference-Native Pretraining Algorithms}
\label{sec:algo_without_optimal_labels}

While \dppt\ demonstrates that preference-only context datasets can support in-context generalization, it does not fully exploit the structure of preference feedback and still relies on optimal action labels during pretraining.
We now introduce two preference-native pretraining frameworks that eliminate the need for both reward supervision and optimal action labels.
These frameworks operate directly on observed preference comparisons and enable pretraining and deployment under preference-only supervision. Full mathematical derivations for the results in this section are provided in Appendix~\ref{sec:app-derivation}. 

\textbf{Learning from Preferences Without an Explicit Oracle.}
As discussed in Section~\ref{sec:icprl}, we treat preferences as primitive supervision signals, and make no assumption about the true process that generates them.
In this section, we introduce learning objectives that convert observed preference comparisons into optimization signals.
We note that these objectives should be understood as surrogate losses that are consistent with preference data, rather than as assumptions about the existence or recoverability of a true reward or advantage function.

\subsection{Overview and Technical Challenges}

Preference-based RL methods typically follow a two-stage paradigm: first infer a latent reward or utility model from preferences, then optimize a policy with respect
to the inferred signal.
However, this paradigm becomes substantially more challenging in ICPRL, where a large number of tasks must be handled simultaneously and each task may provide only
a small number of preference observations.
Naively fitting a separate reward or advantage model per task fails to exploit shared structure across tasks and leads to poor generalization.

Our key insight is to adopt the same principle underlying ICRL: instead of learning task-specific models independently, we train \emph{in-context estimators} that
condition on task-specific context datasets and interpolate across tasks.
We instantiate this idea differently for I-PRL and T-PRL, leading to two distinct frameworks.

\subsection{Pretraining for I-PRL: In-Context Preference Optimization (ICPO)}

Our objective in the I-PRL setting is to pretrain a transformer-based meta-policy $T_\theta(a \mid s, D^I)$ that adapts its behavior based on step-wise preference feedback provided in the context dataset $D^I$.
Unlike task-specific preference optimization, the policy learned here is explicitly conditioned on the context dataset and is expected to infer task structure in context
at deployment time.

To motivate our framework, consider a regularized policy optimization problem defined with respect to a task-dependent scoring function.
If a task-specific score function $A_\tau$ were available, one could define an optimal policy by maximizing expected score subject to a KL constraint with respect to a reference policy, i.e., 
\[
\argmax_{\pi} \bbE_{s}\bigg\{\bbE_{a \sim \pi(a|s)}\left[A_\tau(a,s)\right] - \beta \KL\left(\pi_{\tau}^b(\cdot|s)\|\pi(\cdot|s)\right)\bigg\}, 
\]
where the outer expectation is over states \(s\) in the context dataset \(D\), and \(\pi_\tau^b\) denotes a reference policy for task \(\tau\), which serves as a prior for policy learning. 

In this light, a natural solution is to estimate score functions \(\{A_{\tau_i}(s, a)\}_{i=1}^m\) for all pretraining tasks using a corresponding set of estimators \(\{\widehat{A}_{\tau_i}(s, a)\}_{i=1}^m\), \emph{\color{Black} one for each task}. However, this poses a challenging learning problem. Each pretraining task may have only a limited number of trajectories, and this approach fails to exploit any \emph{\color{Black}shared} structure across tasks that could aid interpolation and generalization.

\textbf{In-context Utility Modeling.}
We address this challenge by introducing an in-context utility proxy $A_\phi(s,a \mid D^I)$, implemented as a transformer that conditions on the preference context dataset.
Crucially, $A_\phi$ does not represent a global or task-invariant advantage function.
Instead, it parameterizes a family of task-specific scoring functions indexed by context, allowing the model to interpolate across pretraining tasks and adapt to new tasks at deployment.

The utility proxy is trained using a surrogate likelihood over observed preference comparisons.
Specifically, we adopt a Bradley--Terry-style objective that encourages consistency between predicted utility differences and observed step-wise preferences:
\begin{equation}
\label{eqn:i-prl-adv-obj}
\begin{aligned}
\max_{\phi} 
\sum_{i,h=1}^{m,H}
\log &\sigma\!\big(
A_\phi(s_{i,h}, a^+_{i,h} \mid D_i^I) -A_\phi(s_{i,h}, a^-_{i,h} \mid D_i^I)
\big),
\end{aligned}
\end{equation}
where $(a^+_{i,h}, a^-_{i,h})$ denotes the preferred and non-preferred actions at step $h$ for task $\tau_i$.
Importantly, $A_\phi$ is not assumed to correspond to a true advantage function; it is a learned utility proxy that induces a preference-consistent ordering.
After training $A_\phi(s,a|D^I)$, we can pretrain the TM policy $T_\theta(a|s,D^I)$ by 
\begin{equation}\label{eqn:i-prl-adv-opt}
\begin{aligned}
      \max_\theta \sum^m_{i=1}\sum^H_{h=1}&\bbE_{a \sim T_\theta(a|s_{i,h},D_i^I)}\left[A_\phi(s_{i,h},a|D_i^I)\right]\\
      &- \beta \KL[\pi_{\tau_i}^b(\cdot|s_{i,h})\|T_\theta(\cdot|s_{i,h},D_i^I)].
\end{aligned}
\end{equation}

{\bf In-Context Preference Optimization.} The above two-step procedure, which first learns $A_\phi(s,a|D^I)$ and then solves~\eqref{eqn:i-prl-adv-opt}, is a complicated procedure. To address this, motivated by the success of direct preference optimization methods for RLHF~\citep{rafailov2023direct}, we observe that the optimization problem in~\eqref{eqn:i-prl-adv-opt} admits a closed-form:
\begin{equation}\label{eqn:i-prl-sol}
    T_{\theta}(a|s,D_i^I) = \frac{\pi^b_{\tau_i}(a|s)\cdot \exp\left(A_\phi(s,a|D_i^I) /\beta\right)}{Z(s,\tau_i)},
\end{equation}
where $Z(s,\tau) = \sum_{a}\pi^b_{\tau_i}(a|s)\exp\left(A_\phi(s,a|D_i^I) /\beta\right)$ is a normalizing constant. In particular, Equation~\eqref{eqn:i-prl-sol} directly relates the policy to the advantage function and motivates a reparameterization of $A_\phi(s,a|D_i^I)$: with some algebraic manipulation on~\eqref{eqn:i-prl-sol}, we have $A_\phi(s,a|D_i^I) = \beta\left(\log T_{\theta}(a|s,D_i^I) + \log Z(s,\tau_i) - \log\pi^b_{\tau_i}(a|s)   \right)$. 
In other words, we can parameterize the in-context advantage estimator with the TM policy $\log T_{\theta}$, and this can be plugged into the learning objective of $A_\phi(s,a|D_i^I)$ in~\eqref{eqn:i-prl-adv-obj} to have
\begin{equation}\label{eqn:i-prl-obj}
    \max_{\theta} \sum_{i,h=1}^{m,H}\log\sigma\left(\beta\cdot\mathcal{J}(a^+_{i,h},a^-_{i,h},s_{i,h}, D^I_i,\pi^b_{\tau_i})\right),
\end{equation}
where $\mathcal{J}(a^+_{i,h},a^-_{i,h},s_{i,h}, D^I_i,\pi^b_{\tau_i})=\frac{\log T_\theta(a^+_{i,h}|s_{i,h}, D^I_i)}{\log\pi^b_{\tau_i}(a^+_{i,h}|s_{i,h})}-\frac{\log T_\theta(a^-_{i,h}|s_{i,h}, D^I_i)}{\log\pi^b_{\tau_i}(a^-_{i,h}|s_{i,h})}$. By this approach, we can directly optimize the TM policy $T_{\theta}(a|s;D^I)$ with a simple supervised learning objective rather than first learning an advantage model and then using RL to solve~\eqref{eqn:i-prl-adv-opt}.

To further simplify the optimization problem, we choose the reference policy $\pi^b_{\tau_i}$ to be the uniformly random policy for all pretraining tasks $\tau_i$, with three main motivations: \textbf{(i)} it leads to consistently strong performance through our experiments in Section~\ref{sec:exp}; \textbf{(ii)} the terms containing $\log\pi^b_{\tau_i}$ in~\eqref{eqn:i-prl-obj} now cancel each other and disappear; \textbf{(iii)} a uniformly random policy motivates exploration, which can be beneficial for deployment to new tasks.
This leads to our final pretraining objective for I-PRL:
\begin{equation}\label{eqn:i-prl-obj-final}
\begin{aligned}
    \max_{\theta} \sum_{i,h=1}^{m,H}\log\sigma\big(\beta\cdot\big(&\log T_\theta(a^+_{i,h}|s_{i,h}, D^I_i)\\
    &-{\color{red}\lambda}\cdot \log T_\theta(a^-_{i,h}|s_{i,h}, D^I_i)\big)\big),
\end{aligned}
\end{equation}
where ${\color{red}\lambda} \in (0,1)$ is a weighting hyperparameter that moderates the influence of non-preferred actions.
The role of \( \lambda \) is critical: it encourages the transformer policy \( T_\theta \) to focus more on increasing the likelihood of preferred actions rather than aggressively suppressing non-preferred ones. Without this adjustment, the model could trivially satisfy the preference constraint by spreading probability mass across many suboptimal actions. 
Our framework can also be applied for \emph{continuous-control} tasks. See Appendix~\ref{sec:app-continuous-control} for details. 

\subsection{Pretraining for T-PRL: In-Context Reward Generation (ICRG)}

In the T-PRL setting, preference feedback is provided only at the trajectory level,
making direct policy optimization less straightforward.
Our strategy is to reduce T-PRL to a standard ICRL problem by generating a dense,
preference-consistent reward signal that can be consumed by existing ICRL pipelines.

\textbf{In-context reward representation.}
We introduce an in-context reward estimator $R_\psi(s,a \mid D^T)$ that conditions on
trajectory-level preference context.
The role of $R_\psi$ is not to recover the true environment reward, but to produce a
reward representation whose induced trajectory ordering matches observed preferences.

We train $R_\psi$ using a surrogate likelihood over trajectory comparisons:
\begin{equation}
\label{equ:preference_eqn}
\begin{aligned}
    \max_{\psi}
\mathcal{L}_R^T(\psi)
=
\sum_{i=1}^M
&\log \sigma\!\bigg(
\sum_{h=1}^H R_\psi(s^+_{i,h}, a^+_{i,h} \mid D_i^T)\\
&-\sum_{h=1}^H R_\psi(s^-_{i,h}, a^-_{i,h} \mid D_i^T)
\bigg),
\end{aligned}
\end{equation}
where $(s_{i,h}^+, a_{i,h}^+)$ is the state-action pair at step $h$ in the preferred trajectory of $D_i^T$ while $(s_{i,h}^-, a_{i,h}^-)$ is the state-action pair in the non-preferred one.
We note that any reward representation that preserves preference ordering is sufficient for our
purposes to extract an optimization signal. 

\textbf{Reward Labeling and Reduction to standard ICRL.}
With a learned reward estimator $R_\psi(s,a|D^T)$, we can label all state-action pairs in the pretraining datasets. Specifically, for each state-action pair $(s_{i,h}, a_{i,h})$ in either the preferred or non-preferred trajectory of $D_i^T$, we append to it an estimated reward $\widehat{r}_{i,h} = R_\psi(s_{i,h},a_{i,h}|D_i^T)$. By this approach, we now have for each pretraining task $\tau_i$ a pretraining dataset containing two trajectories with reward information. 
This enables us to leverage \emph{off-the-shelf} ICRL methods to obtain a pretrained TM policy \( T^R_\theta(a \mid s, D^R) \) that generalizes to new tasks when conditioned on a context dataset \( D^R \). However, as we use an off-the-shelf ICRL method, \( D^R \) is expected to contain reward information. To this end, ICRG uses the learned reward estimator \( R_\psi(s, a \mid D^T) \) to infer such reward signals during deployment, allowing us to deploy \( T^R_\theta(a \mid s, D^R) \) even when the original test-time context datasets contain only preference data.
See Algorithm~\ref{alg:icrg} in Appendix~\ref{sec:app-algo} for pseudocode of the complete pipeline. 
In particular, we choose the \emph{Decision Importance Transformer} framework~\citep{dong2025incontext}, which has the least pretraining data requirements and is specifically designed for pretraining datasets containing suboptimal trajectories. Due to its tangential connection to our contribution, we present it in detail in Appendix~\ref{sec:app-dit}. 

\textbf{Framework Comparisons and Policy Deployments.} To help understand all the proposed ICPRL frameworks, we provide a summary of them with their corresponding models. In addition, we elaborate on their deployments. Due to space constraint, we defer these discussions to Appendix~\ref{app:framework-comparison-and-policy-deployments}.

\section{Experiments}\label{sec:exp}

We empirically demonstrate the efficacy of our proposed frameworks through experiments on various dueling bandit and MDP problems. Due to space constraints, we defer all the bandit experiments to Appendix~\ref{sec:app-bandit-exp} and mainly present the more challenging MDP experiments. See Appendix~\ref{sec:app-implementation} for implementation details and model architectures.

\begin{figure*}[t!]
    \centering
    \begin{subfigure}{\linewidth}
        \centering
        \includegraphics[width=0.90\linewidth]{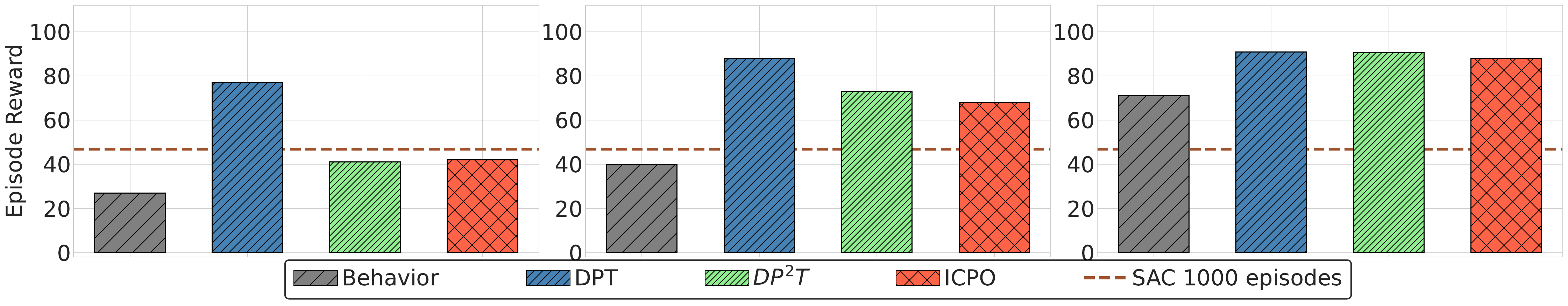}
    \end{subfigure}
    \begin{subfigure}{\linewidth}
        \centering
        \includegraphics[width=0.90\linewidth]{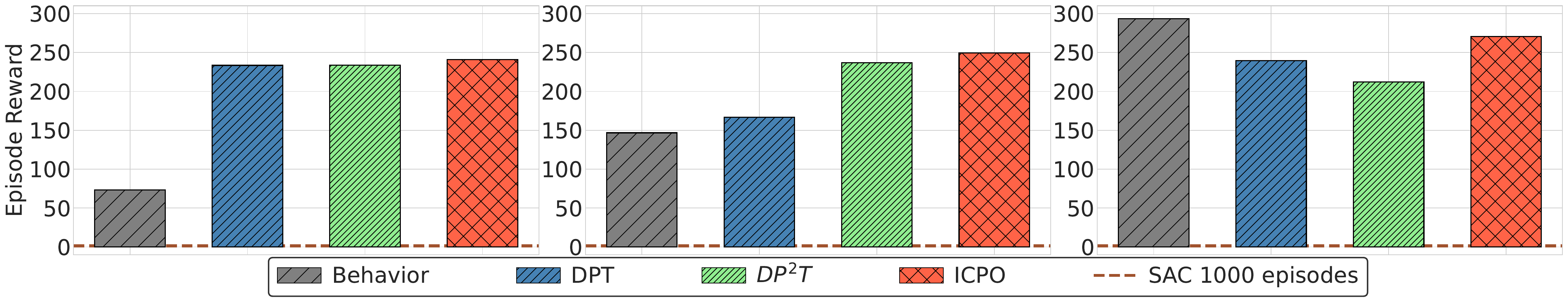}
    \end{subfigure}
    \caption{I-PRL results under context datasets of varying quality (left to right:  {\bf low}, {\bf medium}, and {\bf high} quality) in DarkRoom (top) and Meta-World.}
\end{figure*}

\begin{figure*}
    \centering
    \begin{subfigure}{\linewidth}
        \centering
        \includegraphics[width=0.90\linewidth]{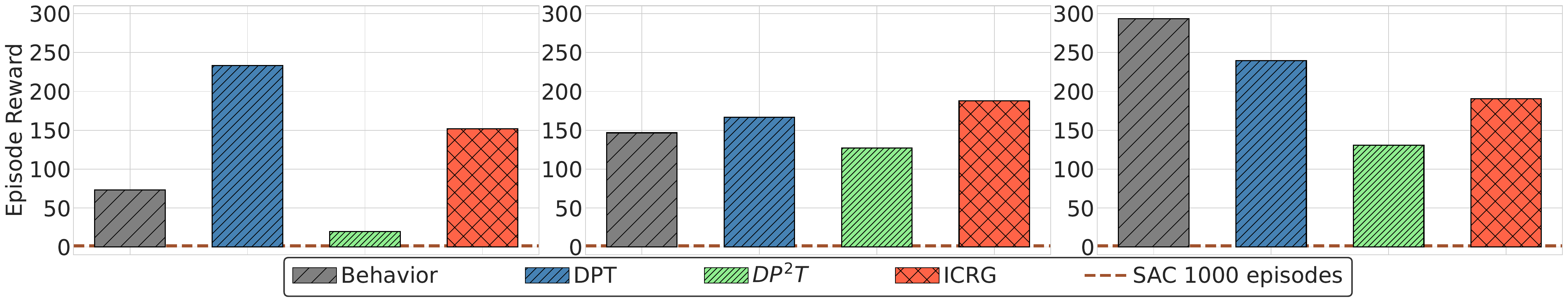}
    \end{subfigure}
\caption{Meta-World (T-PRL) results with context datasets of {\bf low}, {\bf medium}, and {\bf high} quality.}
\end{figure*}

\textbf{Experiments Setup.} We conduct experiments for both I-PRL and T-PRL settings. We consider two representative ICRL tasks: the challenging navigation task {\bf DarkRoom}~\citep{AD} and the complex continuous control task {\bf Meta-World} (reach-v2)~\citep{yu2020meta}. In DarkRoom, the agent is randomly placed in a room with discrete grids. The goal of agent is to move to an \emph{unknown} goal location on one of the grids. The agent has $5$ actions and needs to reach the goal in $H=100$ steps. In Meta-World, the agent controls a robotic hand to reach a target position in $3$D space. We highlight that we use different tasks for pretraining and deployment so that all the test tasks are \emph{unseen} to the ICPRL models. See Appendix~\ref{sec:app-envs} for more details.

\textbf{Synthetic Preference Generation.}
For controlled evaluation and reproducibility, preference labels in our experiments are generated synthetically from the underlying (latent) task reward, which remains
unobservable to the agent.
In the I-PRL setting, step-wise preferences are sampled using a probabilistic comparison model that favors actions with higher optimal advantage at a given state.
In the T-PRL setting, trajectory-level preferences are generated by comparing pairs of trajectories according to their cumulative return.
We use the Bradley--Terry models for both settings, and the precise stochastic models used to generate these preference labels are described
in Appendix~\ref{sec:app-preference-generation}.
We note that this synthetic construction is used solely to instantiate benchmark datasets;
all proposed methods operate exclusively on preference comparisons and never access the latent reward, advantage function, or optimal policy during either pretraining or
deployment.

\textbf{Pretraining Datasets.} To collect pretraining datasets for DarkRoom, we use two behavioral policies: the uniformly random policy and a policy that at every step, with probability $p$ (respectively $1-p$) follows the optimal policy (respectively the uniformly random policy) to choose action. For Meta-World, we construct the pretraining datasets using training checkpoints of \emph{Soft Actor Critic} (SAC). Specifically, SAC is trained until convergence for each task, and we use checkpoints with $\%30$ and $\%80$ performance of the optimal policy as the behavioral policies. 
Due to their different structure, I-PRL and T-PRL follow different data generation processes. See Appendix~\ref{sec:app-pretraining-data} for details.

\textbf{Performance Metric.} We evaluate policies using the standard trajectory cumulative reward. During deployment to a task \(\tau\), we use the ground-truth reward function \(R_\tau(s, a)\), which remains unobservable to the agent, to assign true rewards to all state-action pairs. The total reward is computed by summing these values over the rollout horizon. 

\textbf{Baselines.} To benchmark the performance of our proposed frameworks, we compare against two strong baselines: the state-of-the-art ICRL method \textbf{DPT}; and \textbf{SAC}, a widely used online RL algorithm that trains an agent from scratch in each environment. Implementation details for both baselines are provided in Appendix~\ref{sec:app-baselines}. 
Importantly, comparison to DPT is \emph{\color{Black}inherently unfair}, as it relies on \emph{\color{Black}full reward supervision} and \emph{\color{Black}costly optimal action labels} during pretraining. We include DPT as an \emph{\color{Black}oracle} baseline to contextualize the performance of preference-based methods. To further benchmark the difficulty of the test tasks, we report the performance of SAC trained from scratch for {\color{Black}$1000$} episodes and select the \emph{\color{Black}best} checkpoint for evaluation. In contrast, ICPRL models are deployed to new tasks using at most three offline preference trajectories, underscoring their practicality.








\textbf{Results for I-PRL.} We first observe that both of our ICPRL methods almost always \emph{\color{Black}significantly improve over the behavioral policies} of the context datasets, demonstrating their generalization capabilities. 
For {\bf DarkRoom}, DPT consistently demonstrates the best performance, This is not surprising, as it uses considerably more information for both pretraining and testing. However, when the context dataset has {\bf high-quality}, our ICPRL methods pretrained and deployed \emph{\color{Black}without} any reward information can match DPT. Importantly, ICPO, without using any optimal action labels for pretraining, demonstrates comparable performance to \dppt\ in all cases, showing its \emph{\color{Black}effective usage} of the preference pretraining data. Quite surprisingly, for {\bf Meta-World}, ICPO \emph{\color{Black}consistently outperforms} two supervised pretraining methods, even DPT with full reward supervision. 
We attribute this to the fact that in more complex, continuous domains, action labels are harder to interpret and less informative. As a result, ICPO, which optimizes directly from preference feedback without relying on reward nor action labels, demonstrates superior adaptability.

\textbf{Results for T-PRL.} Due to space constraints, we present the results for the more challenging Meta-World (T-PRL) problem and defer results for DarkRoom (T-PRL) to Appendix~\ref{sec:app-extra-exp}. We first note that, due to sparse (trajectory-level) preference labels, T-PRL is considerably more challenging than I-PRL with step preference. As a result, ICPRL models (ICRG and \dppt\ ) pretrained in T-PRL setting demonstrate less competitive performance than their counterparts in the I-PRL setting.
However, for both {\bf low} and {\bf medium} context datasets, our method ICRG still significantly improves over the behavioral policies. In addition, ICRG consistently outperforms \dppt, showcasing it effectively recovering reward-relevant information from trajectory-level preferences. Most importantly, our ICPRL methods considerably outperform online learning from scratch in the test tasks (SAC with $1000$ episodes), demonstrating their promises towards competitive in-context decision-makers with \emph{\color{Black}completely reward-free pretraining and deployment}.


\section{Discussion and Limitations}\label{sec:diss}
We introduced ICPRL, a new paradigm for training in-context transformer policies without explicit rewards, using only preference supervision. We proposed frameworks for both I-PRL (per-step) and T-PRL (trajectory-level) settings, and showed that pretrained policies generalize effectively across tasks. A key limitation is the assumption of pre-collected offline preference data; future work could explore active preference querying during pretraining to improve data efficiency and generalization.

\section*{Impact Statement}

This paper presents work whose goal is to advance the field of Machine
Learning. There are many potential societal consequences of our work, none
which we feel must be specifically highlighted here.

\bibliography{ref}
\bibliographystyle{icml2026}

\newpage
\appendix
\onecolumn
\section{Related Work}\label{sec:app-related-work}

\textbf{Preference-based Reinforcement Learning (PbRL).}
PbRL focuses on learning from preference signals, typically given as comparisons between actions or trajectories, instead of relying on scalar reward functions~\citep{christiano2017deep, wirth2017survey, brown2019extrapolating,furnkranz2012preference}. Notably, the well-known dueling bandit problem~\cite{yue2012k,wu2016double,dudik2015contextual,komiyama2015regret} is a special case of the PbRL problem without state transitions. 
A common paradigm in PbRL involves a two-stage model-based process: first, learning a reward function from preferences, and second, optimizing a policy using standard RL algorithms with the learned reward~\citep{ibarz2018reward,  pmlr-v139-lee21i, liu2023efficient, liu2022meta}. 
The most salient distinction between the proposed ICPRL paradigm and prior PbRL approaches is that PbRL assumes the \emph{same} environment for training and evaluation, whereas ICPRL operates in the in-context setting, where a model must generalize to \emph{new} tasks at inference time without parameter updates.

\textbf{Offline Reinforcement Learning.} Our work is also related to offline RL\citep{levine2020offline,matsushima2020deployment,prudencio2023survey}. The goal of offline RL is to learn an optimal policy from pre-collected datasets without exploration and interacting with the environments~\citep{wu2019behavior,kidambi2020morel,kumar2020conservative,rashidinejad2021bridging,yin2021towards,jin2021pessimism,dong2023pasta,fujimoto2021minimalist}. In particular, offline RL methods assume that the training data is collected from the environment to be deployed for. To this end, our work falls into the ICRL paradigm, and the goal is to learn a meta-policy from offline preference data collected from diverse RL tasks to generate to unseen RL tasks.

\textbf{Transformer Models and In-Context Reinforcement Learning.} Transformer models have demonstrated superior performance in RL problem~\citep{li2023a,yuan2023transformer}. Building on the in-context learning abilities of transformers, ICRL methods aim to learn a TM-based meta-policy to generalize to unseen tasks in context. ICRL methods differ in their requirements of context datasets. For example, \emph{Algorithm Distillation} ({\bf AD})~\citep{AD} and its variations~\citep{zisman2023emergence,tarasov2025yes} uses sequential modeling to emulate the learning process of RL algorithms, i.e., meta-learning~\citep{vilalta2002perspective}. Decision transformer~\citep{chen2021decision}(DT)-based methods rely on return-to-go to guide the transformer models to generalize to new tasks~\citep{grigsby2023amago,huang2024context,schmied2024retrieval}. A recent framework \emph{Decision Pretrained Transformer} ({\bf DPT}) uses supervised-pretraining for in-context decision making. DPT trains transformers to predict the optimal action given a query state and a set of transitions. Despite its strong performance, DPT requires high-quality action labels for its pretraining. However, to our best knowledge, all existing ICRL methods assume explicit reward (goal) signal for pretraining. To this end, we make the first step towards reward/goal-free ICRL.

\section{Why Supervised-Pretraining Works for ICRL}\label{sec:app-intuition}

As rigorously proved in~\cite{DPT}, the supervised pretraining approach for ICRL can be interpreted as training the TM policy $T_\theta$ to conduct an implicit Posterior Sampling (PS)~\cite{osband2013more}. Under this perspective, during deployment, the decision-making process of a pretrained TM $T_\theta$ is equivalent to conducting three consecutive moves: 
\begin{itemize}
    \item[{\bf (i)}] When given the context dataset $D$, $T_\theta$ first \emph{implicitly} constructs a posterior distribution $p(\tau|D)$ over tasks;
    \item[{\bf (ii)}] $T_\theta$ samples a task $\tau'$ following the constructed posterior $p(\tau|D)$;
    \item[\bf (iii)] $T_\theta$ follows the optimal policy of the sampled task $\tau'$, i.e., $T_\theta(a|s,D) \approx \pi^\star_{\tau'}(s)$. 
\end{itemize}

In particular, similar to other Bayesian approaches, as the size of context dataset $|D|$ increases, the implicit posterior $p(\tau|D)$ concentrates toward the true test task $\tau^\te$, and the sampled task $\tau'$ becomes more similar to $\tau^\te$. 
As a result, if the pretrained TM $T_\theta$ follows the optimal policy for $\tau'$, its performance also increases in $\tau^\te$.

\section{MDP Environment Details}\label{sec:app-envs}
We evaluate our proposed methods on two sequential decision-making environments with contrasting characteristics—one discrete with sparse-reward (DarkRoom) and one continuous with dense-reward (Meta-World Reach-v2)—to assess generalization across task modalities.

\paragraph{DarkRoom.} DarkRoom is a grid-based navigation task with sparse rewards, originally proposed by~\cite{AD}. The environment is a $10 \times 10$ grid, where each cell represents a discrete location the agent can occupy. At the start of each episode, the agent is placed at a random location, and the goal location, unknown to the agent, is randomly chosen and then fixed throughout the episode. The agent can take one of five discrete actions at each time step: up, down, left, right, or no-op. The episode ends either after 100 time steps (the maximum horizon) or upon reaching the goal. A reward of 1 is given only when the agent steps into the goal location; otherwise, all rewards are 0.

We pretrain on 80 randomly sampled tasks (each with a different goal position) and evaluate on a held-out set of 20 tasks. For each task, we generate preference-labeled trajectories using behavior policies that mix optimal and random behavior (see Appendix~\ref{sec:app-pretraining-data}). The discrete action space and binary reward make DarkRoom a useful testbed for evaluating how well our models can leverage sparse preference signals to infer task objectives.

\paragraph{Meta-World (Reach-v2).} Reach-v2 is a continuous control task from the ML1 benchmark of the Meta-World suite~\citep{yu2020meta}. It requires controlling a 7-DoF Sawyer robotic arm to move its end-effector to a 3D target position, which is randomly sampled at the beginning of each task instance. The agent observes both its proprioceptive state (joint positions and velocities) and the target position. The action space is continuous and represents end-effector displacements. At each time step, the agent receives a reward equal to the negative Euclidean distance to the goal. The episode length is fixed at a horizon of 150 time steps.

We use 45 tasks for pretraining and hold out 5 tasks for evaluation. Each task is defined by a different target goal sampled from a bounded region in 3D space. To generate pretraining datasets, we train SAC policies to convergence for each task, and select intermediate checkpoints to simulate suboptimal behavior. Preference labels are generated by comparing trajectories collected from these policies. Due to the continuous nature of the action space and the dense reward function, Reach-v2 is a significantly more challenging setting compared to DarkRoom, providing a strong test of the scalability and adaptability of preference-based models.

{\bf Deployment.} Deployments for both environments are under the offline RL setting: during deployment, no environment interaction is permitted beyond the given offline context. Agents must rely entirely on the offline, preference-labeled context data to infer the task and generalize their behavior.

\section{Baseline Implementation Details}\label{sec:app-baselines}

To benchmark the performance of our proposed ICPRL methods in both the I-PRL and T-PRL settings, we compare against several strong baselines. Each of these baselines represents a distinct paradigm in reinforcement learning: fully supervised transformer-based meta-RL, hybrid preference-supervised methods, and classical model-free reinforcement learning from scratch.

\paragraph{Decision-Pretrained Transformer (DPT).} DPT~\citep{DPT} is a transformer-based meta-RL method that performs in-context adaptation through supervised pretraining. The model learns to predict optimal actions by conditioning on a query state and a reward-labeled trajectory drawn from the same task. During pretraining, DPT is provided with full reward supervision and oracle optimal action labels. For each training task, we use a converged SAC policy to generate reward-annotated context trajectories and extract optimal actions for randomly sampled query states. The model is trained to maximize the log-likelihood of these optimal actions given the context and the query state.

At test time, the pretrained transformer is deployed in a reward-rich offline setting, where it receives reward-annotated context trajectories from unseen tasks and is queried at each step to produce an action. As DPT represents the strongest fully supervised baseline (with access to reward and optimal labels), it serves as an upper bound in our comparison.

\paragraph{DP\textsuperscript{2}T (DPT with Preference-style Context).} This baseline adapts the DPT framework to our reward-free setup by changing the structure of the context data. Instead of using reward-annotated transitions, it consumes preference-labeled trajectories generated from the I-PRL pipeline. However, it retains access to optimal actions for query states during pretraining. In this sense, DP\textsuperscript{2}T is a hybrid method—it leverages the structure of our proposed setting while retaining strong supervision through optimal action labels.

The goal of this baseline is to isolate the effect of removing reward supervision in the context while still keeping access to action supervision. By comparing it against DPT (with full reward and action supervision), and ICPO and ICRG (with neither reward nor action supervision), we can disentangle the individual contributions of reward-rich context and label-rich supervision.

\paragraph{Soft Actor-Critic (SAC).} SAC~\citep{haarnoja2018soft} is a widely used model-free deep RL algorithm that learns stochastic policies in continuous or discrete environments by maximizing a trade-off between expected return and policy entropy. It serves as a strong baseline for traditional reinforcement learning that does not leverage offline in-context data.

In our experiments, SAC plays two roles. First, we use it to train behavioral policies for generating the pretraining datasets in both I-PRL and T-PRL settings. For each pretraining task, we train SAC to convergence and then choose intermediate checkpoints to simulate suboptimal policies (e.g., checkpoints with $20\%$, $40\%$, and $80\%$ of the return of the final converged policy) for context trajectory generation. This allows us to construct preference-labeled data for ICPRL training.

Second, we use SAC as a standalone baseline to measure how well a task-specific policy can perform when trained from scratch. For each test task, we train a SAC agent independently using 1,000 episodes of environment interaction. 
We implement SAC using Stable Baselines3~\citep{stable-baselines3} with default hyperparameters and a fixed episode horizon (100 for DarkRoom, 150 for Meta-World Reach-v2).

Together, these baselines represent a diverse set of RL paradigms and help us contextualize the benefits and trade-offs of purely preference-based pretraining for transformers to generalize to new tasks. 

\section{Synthetic Preference Generation}
\label{sec:app-preference-generation}

We describe the stochastic models used to generate synthetic preference labels for benchmarking.
These models are used \emph{only} to instantiate controlled experimental datasets and are never accessed by the agent during either pretraining or deployment.
All learning algorithms operate exclusively on preference comparisons.

\subsection{Preference Generation for I-PRL}

In the I-PRL setting, preference feedback is provided at the level of individual
decision steps.
For a given task $\tau$, consider a state $s$ and two candidate actions $a$ and $a'$.
We assume the existence of a latent reward function $R_\tau(s,a)$ that defines the
task objective but is not observable to the agent.
Let $\pi_\tau^\star$ denote the optimal policy induced by this reward, and define the
corresponding optimal action-value and value functions as
\[
Q_\tau^\star(s,a) = \mathbb{E}\!\left[\sum_{h=1}^H r_h \,\middle|\, s_1 = s, a_1 = a, \pi_\tau^\star \right],
\quad
V_\tau^\star(s) = \mathbb{E}\!\left[\sum_{h=1}^H r_h \,\middle|\, s_1 = s, \pi_\tau^\star \right].
\]
The optimal advantage function is then given by
\[
A_\tau^\star(s,a) = Q_\tau^\star(s,a) - V_\tau^\star(s).
\]

Given an action pair $(a,a')$ at state $s$, the step-wise preference label
$y \in \{0,1\}$ is sampled according to a Bradley--Terry model:
\[
\mathbb{P}(y = 1 \mid s, a, a', \tau)
=
\mathbb{P}(a \succ a' \mid s, \tau)
=
\sigma\!\left( A_\tau^\star(s,a) - A_\tau^\star(s,a') \right),
\]
where $\sigma(x) = 1 / (1 + \exp(-x))$ is the sigmoid function.
Intuitively, actions with higher optimal advantage are more likely to be preferred,
while allowing for stochasticity in the feedback.

\subsection{Preference Generation for T-PRL}

In the T-PRL setting, preference feedback is provided over entire trajectories.
For a given task $\tau$, let
\[
\xi = \{s_1, a_1, s_2, a_2, \dots, s_H, a_H\},
\quad
\xi' = \{s'_1, a'_1, s'_2, a'_2, \dots, s'_H, a'_H\}
\]
denote two trajectories of horizon $H$, generated by behavioral policies interacting
with the environment.
Trajectories do not contain explicit reward annotations.

Let $R_\tau(s,a)$ denote the latent reward function for task $\tau$.
The preference label $y \in \{0,1\}$ indicating whether $\xi$ is preferred over $\xi'$
is sampled according to a Bradley--Terry model over cumulative return:
\begin{equation}\label{eqn:t-prl-pref}
    \mathbb{P}(y = 1 \mid \xi, \xi', \tau)
=
\mathbb{P}(\xi \succ \xi' \mid \tau)
=
\sigma\!\left(
\sum_{h=1}^H R_\tau(s_h, a_h)
-
\sum_{h=1}^H R_\tau(s'_h, a'_h)
\right).
\end{equation}
Thus, trajectories with higher cumulative latent reward are more likely to be
preferred, while preference feedback remains noisy.

\subsection{Remarks}

The above preference-generation procedures follow standard practice in
preference-based reinforcement learning and are used solely for controlled
evaluation.
The agent never observes $R_\tau$, $Q_\tau^\star$, $V_\tau^\star$, $A_\tau^\star$, or
any optimal policy.
All proposed methods receive only preference comparisons as input and must infer
task structure through in-context learning.

\section{Pretraining Data Generation}\label{sec:app-pretraining-data}

We construct our pretraining datasets to simulate realistic, reward-free supervision via preferences. Our approach leverages varying-quality behavioral policies to provide diverse experiences across tasks. We separately describe the data generation process for each environment and for both the I-PRL and T-PRL settings.

\paragraph{DarkRoom.} 
To construct the context dataset, we generate trajectories using a \textbf{mixed policy} that interpolates between the optimal and random policies. Specifically, at every time step, the action is chosen following the optimal policy with probability $p$, and with probability $1-p$ following a uniformly random policy. We adjust $p$ to control the overall trajectory return, ensuring the resulting policies yield approximately 20\%, 40\%, or 80\% of the optimal cumulative reward. This allows us to evaluate how the quality of context dataset impacts generalization.

For the \textbf{I-PRL} setting, at each time step $h$, we sample a pair of actions, one from the uniformly random policy and the other from a mixed policy defined as above. 
Then we compute the preference labels using the (optimal) advantage function. 
We note that the optimal advantage function for DarkRoom has a closed form. After the preference label is generated, the current state transits according to the preferred action. We collect $1000$ trajectories for each task. 

In the \textbf{T-PRL} setting, to construct a pair of trajectories, we rollout twice a mixed policy defined above to have two trajectories. Then the trajectory preference labels are generated following the BT model defined in Equation~\eqref{eqn:t-prl-pref}. We note that the true reward functions are only for preference label generation, and they are not accessible during pretraining. For each task, we repeat this process to have $5000$ trajectory pairs and their associated preference labels. 

\paragraph{Meta-World (Reach-v2).} Each task in the Meta-World ML1 Reach benchmark is defined by a unique 3D goal position for the robotic arm. We train a separate SAC agent per task to convergence, then select multiple checkpoints to serve as behavioral policies with different proficiency levels. Specifically, we extract checkpoints that achieve approximately 20\%, 40\%, and 80\% of the final SAC return and use them to generate demonstration trajectories. These trajectories vary in quality and coverage, offering a broad distribution of behaviors for preference comparisons.

In the \textbf{I-PRL} setup, at each time step $h$, we sample a pair of actions, one from the checkpoint with $20\%$ of the final return and the other from the checkpoint with $80\%$ of the final return. Then we compute the preference labels using the (optimal) advantage function. Since the optimal advantage function for Meta-World is unknown, we approximate it using the value functions learned by converged SAC policies. Let $\widehat{Q}(s,a)$ be the learned action-value function of a converge SAC policy. We approximate the optimal advantage function with $A^\star(s,a) \approx \widehat{Q}(s,a) - \bbE_{a\sim \pi}[\widehat{Q}(s,a)]$ where $\pi$ is the converged SAC policy and we estimate the expectation with Monte Carlo simulation, i.e., sampling a lot of actions from $\pi$ and average their $\widehat{Q}(s,a)$ values. After the preference label is generated, the current state transits according to the preferred action. We collect $1000$ trajectories for each task. 
 

In the \textbf{T-PRL} setup, we first sample two trajectories with the SAC checkpoints with $50\%$ return of the converged policies. Then the trajectory preference labels are generated following the BT model defined in Equation~\eqref{eqn:t-prl-pref}. We note that we only use the true reward functions during preference label generation, these reward functions are unobservable during pretraining. For each task, we repeat this process to have $5000$ trajectory pairs and their associated preference labels.



\section{Implementation Details}\label{sec:app-implementation}
All transformer models use the GPT-2 architecture~\citep{gpt2}. We adopt causal masking and position embeddings as in standard autoregressive transformers. For all experiments in the I-PRL setting, we use transformers with $6$ layers, $256$ hidden dimensions, and $8$ attention heads. For experiments in the T-PRL setting, we use transformer models with $4$ layers, $32$ hidden dimensions, and $4$ attention heads for DarkRoom and models with $8$ layers, $256$ hidden dimensions, and $8$ attention heads for Meta-World.
For both policy models $T_\theta$ and reward estimators $R_\psi$, we format the input sequence as tokenized tuples of states, actions, preferences, next states, and/or rewards depending on the setting.  

\begin{figure}[h]
\centering
\includegraphics[width=0.90\linewidth]{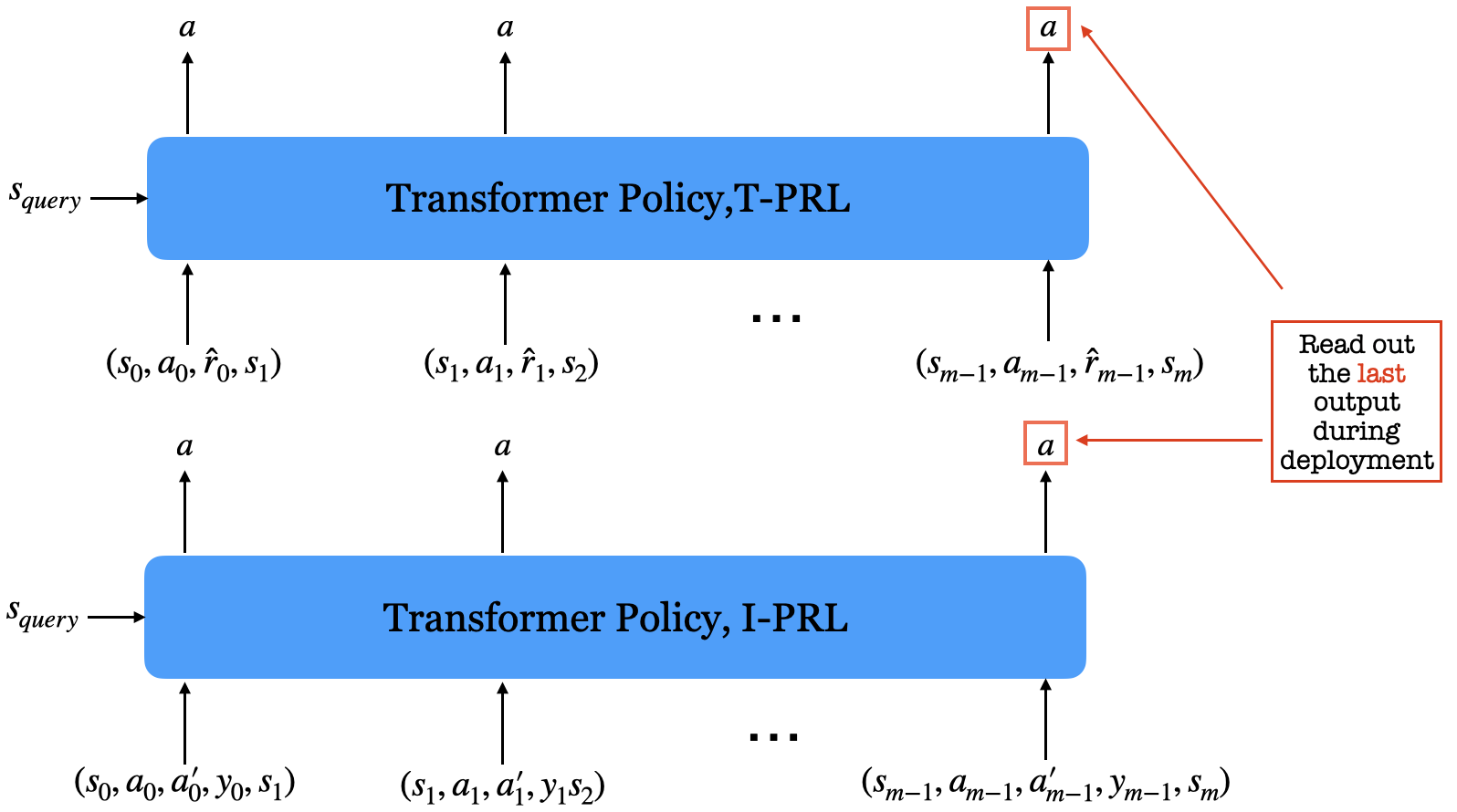}
\caption{Transformer policy architecture. The {\bf bottom} model illustrates the architecture of TM policies for the I-PRL setting. The {\bf top} depicts the T-PRL setting where the reward values $\widehat{r}_h$ are approximated by the in-context reward estimator.}
\label{fig:transformer_policy}
\end{figure}

\paragraph{Transformer Policy Architecture.}Figure~\ref{fig:transformer_policy} shows the architecture of the TM policy $T_\theta$ for both the T-PRL and I-PRL settings.

\begin{figure}[h]
    \centering
    \includegraphics[width=0.80\linewidth]{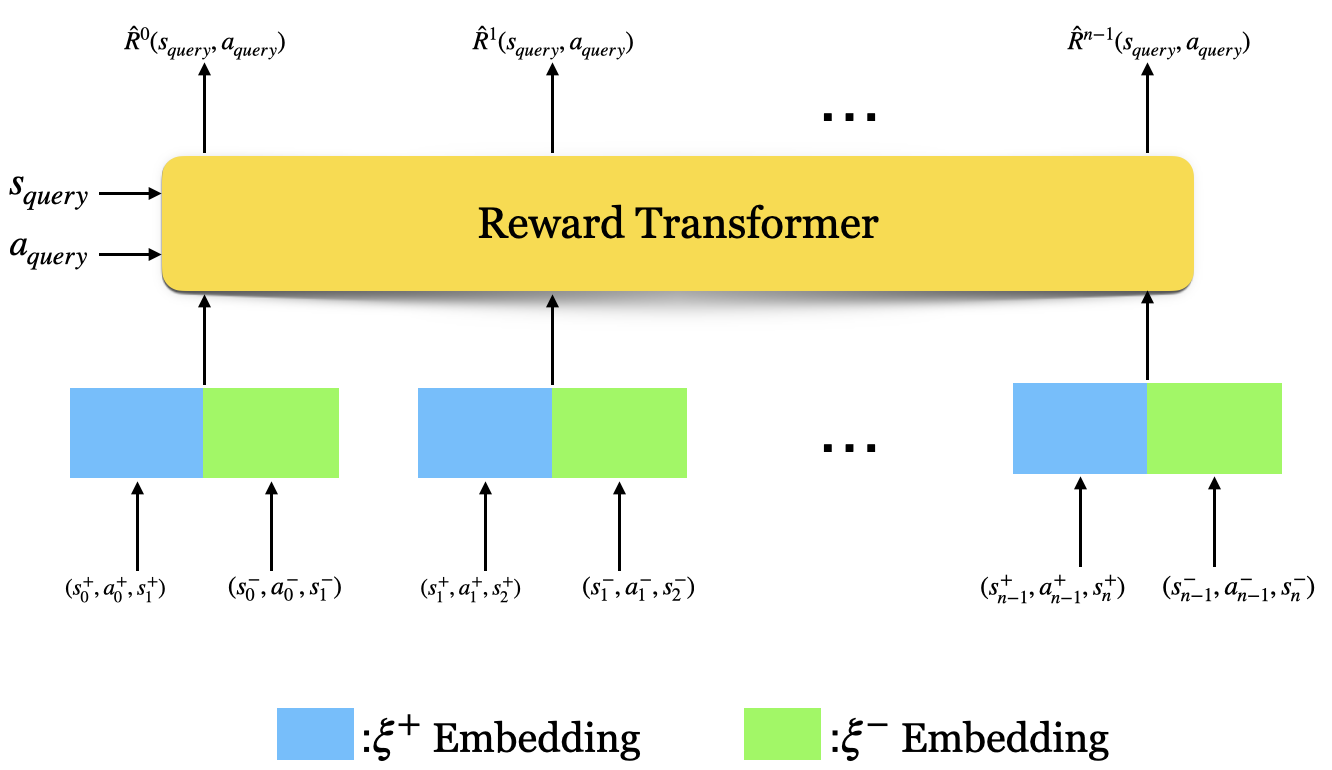}
    \caption{The architecture of the reward transformer.}
    \label{fig:icrg_arch}
\end{figure}

\paragraph{Reward Model Architecture.} 
For the in-context reward estimator $R_\phi(s,a|D^T)$, we use two separate modules to independently embed the transitions from the preferred and non-preferred trajectories, respectively denoted as $\xi^+$ and $\xi^-$. Specifically, conditioned on the context dataset $(\xi^+,\xi^-)$, $R_\phi(s,a|D^T)$ takes as input a query state-action pair and outputs an estimated reward for the given query state-action pair.  
Figure~\ref{fig:icrg_arch} illustrates its design and model architecture. When using $R_\phi(s,a|D^T)$ to estimate rewards, either to pretrain the DIT model or during deployment, we read out its prediction at the last time step as the estimated reward. 

\paragraph{Optimization.} All models are trained using the Adam optimizer with a learning rate of $3 \times 10^{-4}$ and batch size of 32. We use early stopping based on validation performance on a small hold-out pretraining dataset.

\subsection{ICPO for Continuous Control Tasks}\label{sec:app-continuous-control}
{\bf Continuous Control.} In the continuous control setting, we model the policy as a Gaussian distribution \(\mathcal{N}(a; T_\theta(s, D^I), \gamma \mathbf{I})\), where the transformer-based policy \(T_\theta\) predicts the mean and \(\gamma\) controls the fixed diagonal covariance. Under this formulation, the pretraining objective simplifies to:
\begin{equation}\label{eqn:i-prl-continuous}
    \frac{1}{MH} \sum_{i=1}^M \sum_{h=1}^H \log \left( \sigma \left( \beta \cdot \left( \| a_{i,h}^- - T_\theta(s_{i,h}, D^I_i) \|_2 - \| a_{i,h}^+ - T_\theta(s_{i,h}, D^I_i) \|_2 \right) \right) \right),
\end{equation}
where \(a_{i,h}^+\) and \(a_{i,h}^-\) denote the preferred and less-preferred actions at step \(h\) in task \(\tau_i\). See its full derivation in Section~\ref{sec:app-derivation}.

\section{Framework Comparisons and Policy Deployments}\label{app:framework-comparison-and-policy-deployments}
We summarize the ICPRL frameworks introduced so far along with their corresponding models: 
\textbf{(a) $\mathbf{DP^2T}$} yields task-conditioned policies {\color{Black}\(T^S_\theta(a|s, D^I)\)} for I-PRL and {\color{Black}\(T^S_\theta(a|s, D^T)\)} for T-PRL, conditioning on different context datasets; 
\textbf{(b) ICPO} (I-PRL only) produces a similar policy {\color{Black}\(T^I_\theta(a \mid s, D^I)\)} with the same inference structure as \dppt\, but differs in pretraining: it does not require optimal action labels and is trained with a preference-based supervised learning objective.
\textbf{(c) ICRG} (T-PRL only) outputs an in-context reward estimator {\color{Black}\(R_\psi(s,a|D^T)\)} and an ICRL policy {\color{Black}\(T^R_\theta(a|s, D^R)\)} requiring reward-labeled context data.

\textbf{Deployment Setup.} Due to the different structures of I-PRL and T-PRL, these models have distinct deployment procedures. In both settings, we first sample an unseen test task \(\tau^{\te} \sim p_\tau\). 

\textbf{$\mathbf{DP^2T}$ Deployment.} In both I-PRL and T-PRL, \dppt\ assumes access to a context dataset \(D_{\tau^{\te}}\) collected from an unknown behavior policy in the test task $\tau^{\te}$. This dataset corresponds to \(D^I_{\tau^{\te}}\) in I-PRL and \(D^T_{\tau^{\te}}\) in T-PRL. At each time step \(h\), the agent observes state \(s_h\) and samples an action from the TM policy \(a_h \sim T^S_\theta(a_h|s_h, D_{\tau^\te})\). 

{\bf ICPO Deployment (I-PRL).} Similar to \dppt, ICPO assumes access to a context dataset \(D^I_{\tau^{\te}}\). At each step \(h\), the agent acts according to the TM policy $a_h \sim T^I_\theta(a_h|s_h, D^I_{\tau^{\te}})$.

{\bf ICRG Deployment (T-PRL).} A context dataset $D_{\tau^\te}^T$ is first sampled. Subsequently, a behavior trajectory with only state-action pairs $D=\{s_1,a_1,s_2,a_2,\dots,s_H,a_H\}$ is sampled from $\tau^\te$. Using the pretrained in-context reward estimator \(R_\psi\), each state-action pair \((s_h, a_h)\) is annotated with a predicted reward \(\widehat{r}_h = R_\psi(s_h, a_h|D^T_{\tau^{\te}})\). The resulting reward-augmented context \(D^A = \{s_h, a_h, \widehat{r}_h\}_{h=1}^H\) is used to prompt the ICRL policy $a_h \sim T^R_\theta(a_h|s_h, D^A)$ to take actions $a_h$ at each step $h$. 

\section{Derivations}\label{sec:app-derivation}

\subsection{Derivation of Equation~\eqref{eqn:i-prl-sol}}

For any fixed task $\tau$, state $s$, context dataset $D^I$ and reference policy $\pi_\tau^b$ for $\tau$, we have
\begin{align*}
 &\max_{\pi}\bbE_{a \sim \pi(a|s;D^I)}\left[A_{\phi}\left(s,a|D^I\right) - \beta\cdot\KL(\pi(\cdot|s;D^I) \| \pi_{\tau}^b(\cdot|s))\right] \\
 &= \min_{\pi}\bbE_{a \sim \pi(a|s;D^I)}\left[\log\frac{\pi(a|s;D^I)}{\pi_{\tau}^b(a|s)}-\frac{1}{\beta}A_{\phi}\left(s,a|D^I\right)\right]\\
 &= \min_{\pi}\bbE_{a \sim \pi(a|s;D^I)}\left[\log\frac{\pi(a|s;D^I)}{\pi_{\tau}^b(a|s)\exp(A_{\phi}\left(s,a|D^I\right)/\beta)}\right] \\
 &=\min_{\pi}\bbE_{a \sim \pi(a|s;D^I)}\left[\log\frac{\pi(a|s;D^I)}{\pi_{\tau}^b(a|s)\exp(A_{\phi}\left(s,a|D^I\right)/\beta)/Z(s,\tau)}-\log Z(s,\tau)\right] \\
  &= \min_{\pi}\bbE_{a \sim \pi(a|s;D^I)}\left[\log\frac{\pi(a|s;D^I)}{\pi_{\tau}^b(a|s)\exp(A_{\phi}\left(s,a|D^I\right)/\beta)/Z(s,\tau)}\right]\quad\text{($Z(s,\tau)$ is independent of $\pi$)} \\
   & = \min_{\pi}\KL(\pi(\cdot|s;\tau) \| \pi_{\tau}^\star), 
\end{align*}
where $\pi_{\tau}^\star(a|s) = \pi_{\tau}^b(a|s)\exp(A_{\phi}\left(s,a|D^I\right)/\beta)/Z(s,\tau)$. Note that the optimum $\pi$ for a fixed $s$ and task $\tau$ is obtained at $\pi = \pi_{\tau}^\star$, which is unique by the uniqueness property of KL divergence, i.e., $\KL(\pi \| \pi_{\tau}^\star) = 0$ if and only if $\pi =\pi_{\tau}^\star(a|s)$.

\subsection{Derivation of Equation~\eqref{eqn:i-prl-continuous}}
In the case of continuous case, with a notation overload, we choose $T_\theta(a|s;D^I) \sim \cN(a;T_\theta(s,D^I),\gamma \mathbf{I})$ such that we use the TM policy to directly predict the mean action $T_\theta(s,D^I) \in \bbR^d$ where $d$ is the action dimension. Thus, we have 
\begin{align*}
    \log T_\theta(a|s, D_i) = \log C(d,\gamma) -\frac{1}{2\gamma}\|a-T_\theta(s,D^I)\|_2^2,
\end{align*}
where $C(d,\gamma)$ is a constant only involving universal constants, $d$ and $\gamma$. With this identity, for any state $s$ and pair of actions $a^+$ and $a^-$, 
\begin{align*}
 &\log\sigma\left(\beta\cdot\left(\log T_\theta(a^+|s, D^I)-\lambda\cdot \log T_\theta(a^-|s, D^I)\right)\right) \\
 &=\log\sigma\left(\beta\left(\log C(d,\gamma) -\frac{1}{2\gamma}\|a^+ -T_\theta(s,D^I)\|_2^2 - \log C(d,\gamma) +\frac{1}{2\gamma}\|a^- -T_\theta(s,D^I)\|_2^2\right)\right) \\
 &= \log \sigma\left(\beta\cdot \left( \| a^- - T_\theta(s, D^I) \|^2_2 - \| a^+ - T_\theta(s, D^I) \|^2_2 \right) \right).
\end{align*}

\section{Decision Importance Transformer}\label{sec:app-dit}
The \emph{Decision Importance Transformer} (DIT) framework addresses the challenge of obtaining optimal action labels for query states during pretraining~\citep{dong2025incontext}. We use DIT as a module for our proposed framework for the T-PRL setting, specifically for pretraining without query states and optimal action labels. During pretraining, DIT requires for each pretraining task $\tau_i$ a context dataset $D_i=\{\xi_i\}$ containing a trajectory $\xi_i$ with reward information. For simplicity, we assume that each context dataset only contains one trajectory. 

To apply DIT in our ICPRL setting, we first pretrain an in-context reward estimator $R(s,a|D^I)$ and then use it to label all the trajectory pairs in the context dataset. After this step, we have trajectories with no missing reward information (all the rewards are estimated by $R(s,a|D^I)$), and we use DIT to pretrain a TM policy that can generalize to new tasks. In particular, we also rely on $R(s,a|D^I)$ to generate all the required reward information during deployment so that the DIT pretrained models, which require reward information in the context dataset, can be deployed in new tasks. Next we discuss DIT in detail.  

\begin{figure}[htbp]
    \centering
    \begin{subfigure}[t]{0.45\textwidth}
        \centering
        \includegraphics[width=\linewidth]{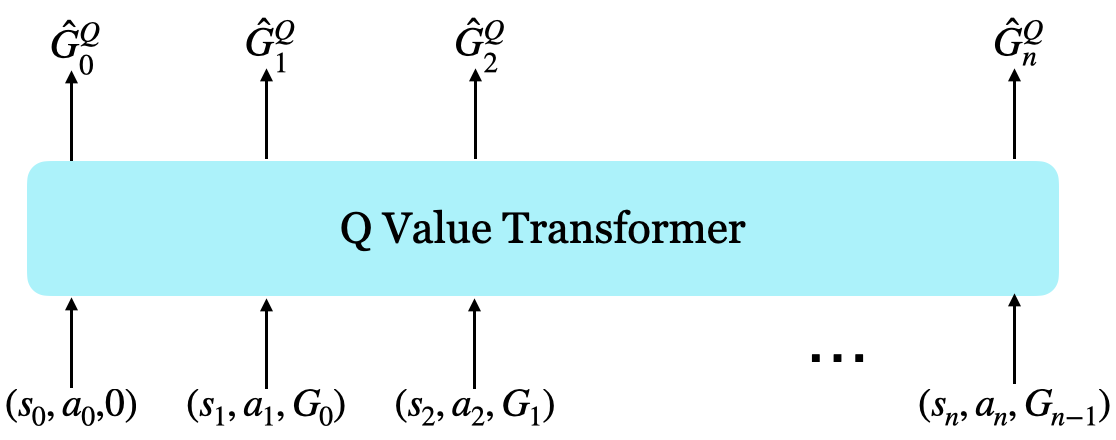}
        \caption{The Architecture of the Q-value transformer.}
        \label{fig:QValueTransformer}
    \end{subfigure} \hfill
    \begin{subfigure}[t]{0.45\textwidth}
        \centering
        \includegraphics[width=\linewidth]{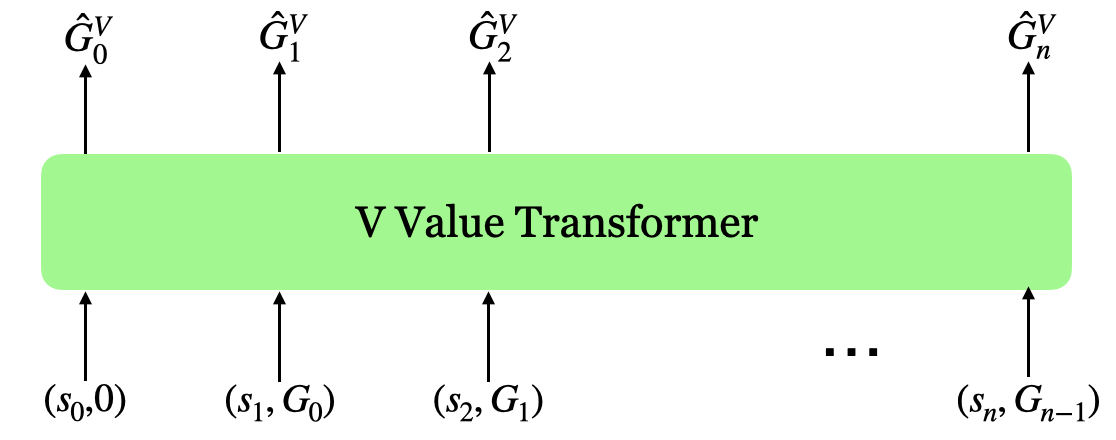}
        \caption{The Architecture of the V-value transformer.}
        \label{fig:VValueTransformer}
    \end{subfigure}
    \caption{Overview of the DIT architecture. The transformers approximate the Q and V values separately, which are then used to compute the advantage of a specific state-action pair.}
    \label{fig:DiT_Arch}
\end{figure}

\paragraph{Optimization Objective.}DIT approximates the advantage of a state-action pair by fitting two transformers in parallel. Let $G^i_h = \sum_{h'-h}^H\gamma^{h'-h}r_h^i$ denote the in-trajectory discounted cumulative reward for the trajectory $\xi_i$ with reward information in the context dataset $D_i$ for task $\tau_i$. For each state-action pair,  $\hat{Q}_{\zeta}(s^i_h, a^i_h|D_Q^{h,i})$ and $\hat{V}_{\psi}(s^i_h|D_V^{h,i})$ estimate the action-value and state-value functions, respectively. We compute the discounted cumulative rewards at each step of a trajectory and construct training datasets of the form $D^i_Q= \{(s^i_h, a^i_h, G^i_h)\}_{h=1}^{H-1}$ and $D^i_V= \{(s^i_h,  G^i_h)\}_{h=1}^{H-1}$ for every trajectory. The V-value transformer takes a sequence of states as input to estimate state values, while the Q-value transformer processes state-action pairs to approximate Q-values. Figure \ref{fig:DiT_Arch} illustrates the two transformers, each processing a different input sequence. We train $\hat{Q}_{\zeta}$ and $\hat{Q}_{\zeta}$ with the following objective function: 
\begin{equation}\label{eqn:dit-loss}
    \operatorname*{min}_{\zeta, \theta} \sum_{i=1}^m \sum_{h=1}^H (\hat{Q}_{\zeta}(s^i_h, a^i_h|D_Q^{h, i}) - G^i_h)^2 + (\hat{V}_{\psi}(s^i_h|D_Q^{h, i}) - G^i_h),
\end{equation}
where $m$ denotes the total number of trajectories in the pretraining dataset, and $D_Q^{h,i}$ and $D_V^{h, i}$ contains the first $h$ tuples of $D_Q^{i}$ and $D_V^{i}$.
After training, we calculate the advantage value of a state-action pair for task $\tau_i$ as
\begin{equation}\label{eqn:dit-advantage}
    \hat{A}_b(s^i_h, a^i_h|\tau_i) = \hat{Q}_{\zeta}(s^i_h, a^i_h|D_Q^{h,i}) -  \hat{V}_{\phi}(s^i_h|D_V^{h, i}).
\end{equation}

$\hat{A}(s^i_h, a^i_h|\tau^i)$ works as an approximation to the true advantage function, we now could use it for the optimization of the policy transformer $T^R_{\theta}$ as
\begin{equation}\label{eqn:dit-optimization}
    \theta^{\star} \in \operatorname*{argmin}_{\theta \in \Theta} - \frac{1}{mH}\sum_{i=1}^m\sum_{h=1}^H \text{exp}(\hat{A}_b(s_h^i, a_h^i|\tau_i)/\eta )\log T^R_{\theta}(a_h^i|s_h^i, D^i).
\end{equation}
Intuitively, for a trajectory $\tau_i$ from a task, DIT optimizes the TM policy with a weighted supervised pretraining objective where the weights are calculated through the advantage function. In particular, it assigns more weights for better actions, i.e., actions with higher advantage values.  

\paragraph{Implementation of DIT's Value Transformers.} For DarkRoom, we use a GPT-2 architecture with 4 layers, 32 hidden dimensions, and 4 attention heads, as originally reported by~\citep{DPT}. Meta-World (Reach-v2) requires a larger model, so we adopt a GPT-2 with 8 layers, 256 hidden dimensions, and 8 attention heads. Algorithm~\ref{algo:DIT} provides the pseudocode for training DIT.

\begin{algorithm}
\caption{Pretraining of Decision Importance Transformer}\label{algo:DIT}
\begin{algorithmic}[1] 
\STATE \textbf{Input:} Pretraining Dataset $\cD=\{D^i\}$; transformer models $T_\theta, \widehat{Q}_{\zeta}, \widehat{V}_{\phi}$.
\STATE {\color{red}\fontfamily{cmtt}\selectfont // \; In-context Estimation of Advantage Functions}
\STATE{Randomly initialize and train $\widehat{Q}_{\zeta}$ and $\widehat{V}_{\phi}$ by optimizing the loss in Equation~(\ref{eqn:dit-loss}).}
\STATE{Construct the in-context advantage estimator as:
$$
\widehat{A}_{b}=\widehat{Q}_{\zeta}-\widehat{V}_{\phi}.
$$}
\STATE {\color{red}\fontfamily{cmtt}\selectfont // \; Weighted Pretraining}
\STATE{Randomly initialize $T_\theta$.}
\STATE{With trained $\widehat{A}_{b}$ and $\cD$, train $T_{\theta}$ by optimizing the loss in Equation~(\ref{eqn:dit-optimization}).}
\end{algorithmic}
\end{algorithm}

\section{Extra Experiments}\label{sec:app-extra-exp}
\subsection{Dueling Bandit Experiments}\label{sec:app-bandit-exp}
We consider dueling bandit ({\bf DB}) problems with a shared linear bandit structure across varying DB problems~\cite{yue2012k}. 
All DB problems have the same action space $\cA$, i.e., same number of bandits. To facilitate generalization to new DB problems, we assume a \emph{fixed} bandit feature mapping $\phi:\cA \rightarrow \bbR^d$ such that $\phi(a) \in \bbR^d$ is the feature for bandit $a$ shared by all DB problems. Each DB problem $\tau$ is characterized by a vector $\theta_\tau \in \bbR^d$ such that the expected reward of bandit $a$ for task $\tau$ is defined as $r_{\tau}(a) = \theta^{\intercal}_{\tau}\phi(a)$. To create challenging bandit problems with stochasticity, we assume that the observed reward is stochastic, that is, the observed reward after selecting $a$ in task $\tau$ follows a Gaussian distribution $r \sim \cN(r_{\tau}(a),\kappa^2)$ where $\kappa^2 = 0.3$ represents variance of reward observations. 

In a DB problem, at each time step $h \in [H]$ within a horizon $H$ , the agent chooses two actions $a,a'$ and receives a preference label $y$ on which of the two chosen actions is more preferred. 
Specifically, we follow the BT model to assume that $\bbP(y=1|a,a',\tau) = \bbP(a\succ a|\tau) = \sigma(r_\tau(a)-r_\tau(a'))$. 
We underscore that $r_\tau$ is not observed in DB problems and we need to infer its information solely from the preference label $y$. The {\bf goal} of DB problems is to find a bandit $a^\star \in \argmax_a r_\tau(a)$ that maximizes the expected reward. This is equivalent to a von Neumann winner which has probability at least $0.5$ to be preferred over any bandits. We choose $|\cA|=20$, $d=10$, and $H=200$. 

Note that the DB problems are special cases of our proposed {\bf I-PRL} setting without state transitions. To this end, we evaluate our proposed {\bf ICPO} framework on DB problems, as \emph{\color{Black}it is designed for the I-PRL setting} and \emph{\color{Black}its pretraining does not require the optimal bandit information}. The pretraining dataset for ICPO are generated as follows. 

\textbf{Pretraining Dataset.} We first generate the bandit features $\phi(a)\sim \sim \cN_d\left(0,I_d/d\right), \forall a \in \cA$, independently following a Gaussian distribution. We generate $m$ DB problems in total and one context dataset for each generated DB problem. To this end, for each pretraining DB problem $\tau_i$, we sample its parameter $\theta_i$ independently from other problem parameters following $\theta^i \sim \cN_d\left(0,I_d/d\right)$. To generate the context dataset $D^I$ for $\tau_i$, at each step $h$, we randomly sample a pair of distinct actions $(a_h,a'_h)$ following a uniform distribution over all pairs of different bandits. We do not enforce any extra coverage of the optimal bandits. We collect $100$k context datasets for DB problems.

\textbf{Evaluation and Baselines.} To benchmark our ICPRL frameworks, we compare them to \emph{Double Thompson Sampling} ({\bf DTS})~\cite{wu2016double}, one of the most competitive DB algorithms. 
We deploy the pretrained TM policy $T_\theta$ and DTS to \emph{\color{Black}new} DB problems. 
In the offline setting, we are given a context dataset $D^I$. Conditioned on $D^I$, our method ICPO follows the policy $T_\theta(a|D^I)$ to choose the first action $a \in \argmax_{\Tilde{a}} T_\theta(\Tilde{a}|D^I)$ and randomly sample the second action $a'$; DTS also utilizes the same bandit preferences in $D^I$ to select a pair of actions $(a,a')$, as detailed in~\cite{wu2016double}.
In terms of metrics, we follow the convention to use the \emph{weak regret} defined as follows. For a given pair of actions $a$ and $a'$, their weak regret $\mathrm{reg}$ for a DB problem $\tau$ is $\mathrm{reg}(a,a') = \min\left(R_\tau^\star-R_\tau(a),R_\tau^\star-R_\tau(a')\right)$,
where $R^\star_\tau$ is the optimal expected reward for $\tau$. 
We evaluate both ICPO and DTS from three types of context data with different types of behavioral policies: (i) uniformly random behavioral policies; (ii) behavioral policies sampled from a Dirichlet distribution with uniform prior; (iii) DTS.

\begin{figure}[H]
    \centering
    \includegraphics[width=1.0\linewidth]{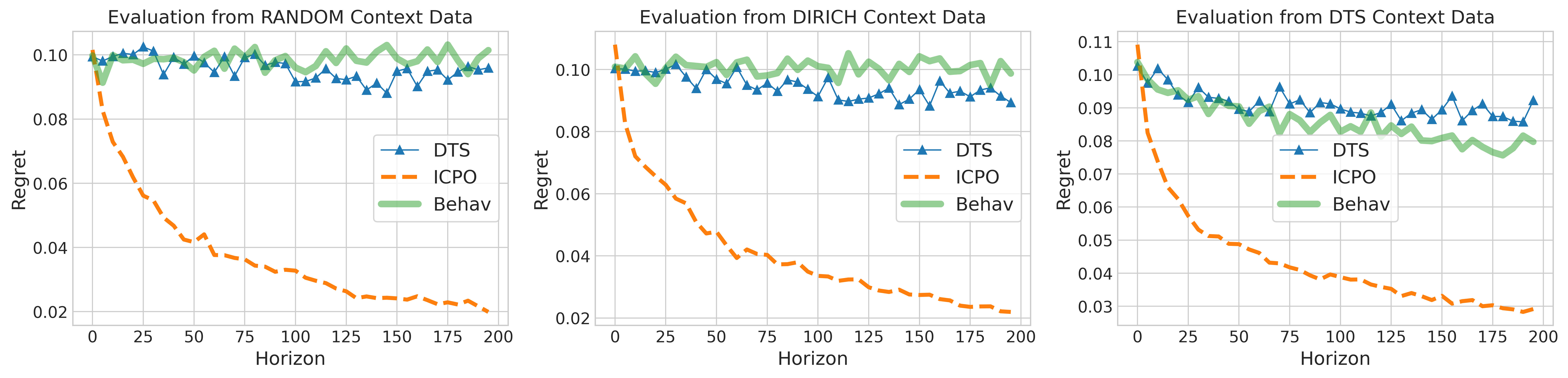}
    \caption{Performance Comparison for Dueling Bandit Problems. We evaluate performance under three types of context datasets, generated by different behavioral policies. {\bf Random} context data is generated by a uniformly random behavioral policy. {\bf Dirichlet} context data is generated by a behavioral policy sampled from the uniform Dirichlet distribution. {\bf DTS} context data is generated by the competitive DB algorithm Double Thompson Sampling.}
    \label{fig:db-online}
\end{figure}

\textbf{Results Discussion.} We present key results in Figures~\ref{fig:db-online}. Note that the DB problems we consider are indeed challenging problems because the agent needs to identify the optimal bandit out of $|\cA|=20$ bandits within only $H=200$ steps. This is verified in Figures~\ref{fig:db-online}, where DTS struggles to reduce its regret. Despite of these challenges, with all three context data, \emph{\color{Black}our method ICPO significantly outperforms DTS}, showing considerably \emph{\color{Black}faster regret decrease} and \emph{\color{Black}consistently improving over the behavioral policies}. These results prove that our framework can pretrain transformers to efficiently solve new DB problems. In addition, the pretrained TM policies also demonstrate \emph{\color{Black}robustness to distribution shift of context data}: although the pretraining data are generated by the uniformly random policies, ICPO consistently demonstrate strong performance even when the context data comes from Dirichlet behavioral policies or DTS.

\subsection{MDP Experiments}




\textbf{T-PRL results in Darkroom.}
\begin{figure}[H]
    \centering
    \begin{subfigure}{\linewidth}
        \centering
        \includegraphics[width=\linewidth]{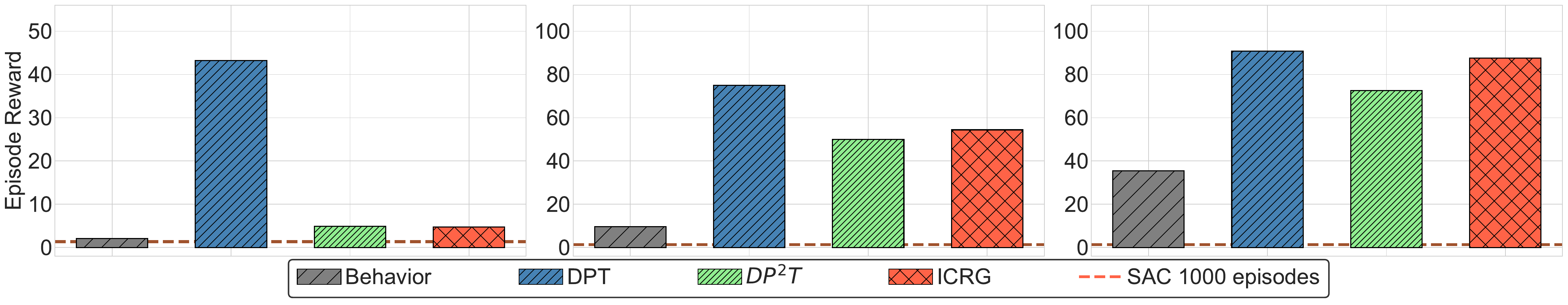}
        \caption{Darkroom (T-PRL) \textbf{average episode reward} with context datasets of {\bf low}, {\bf medium}, and {\bf high} quality.}        
    \end{subfigure}
\end{figure}


\section{Impact of $\lambda$ for ICPO}

In this section, we analyze the effect of  the hyperparameter \(\lambda\) on the episode rewards. 

We conduct two different ablation experiments in DarkRoom and Meta-World to comprehensively understand the effect of $\lambda$. In Darkroom, we use $\lambda$ to scale the log-probability of \emph{\color{Black}non-preferred} actions while in Meta-World we use $\lambda$ to scale the log-probability of \emph{\color{Black}preferred} actions.  When {\bf scaling non-preferred actions}, smaller values of $\lambda$ (e.g., $\lambda <1$) motivate the TM policy to focus more on \emph{\color{Black}increasing the log-probability of preferred actions}. Thus, the model performance is expected to increase with when $\lambda$ decreases. In contrary, when {\bf scaling preferred actions}, a small $\lambda$ (e.g., $\lambda <1$) in fact motivates the TM policy to focus on \emph{\color{Black}decreasing the log-probability of non-preferred actions}. However, this does not guarantee to increase the log-probability of preferred actions. Thus, when using $\lambda$ to scale preferred actions, we should choose large values for $\lambda$, e.g., $\lambda >1$.  

\begin{figure}[H]
    \centering
    \begin{subfigure}{\linewidth}
        \centering
        \includegraphics[width=\linewidth]{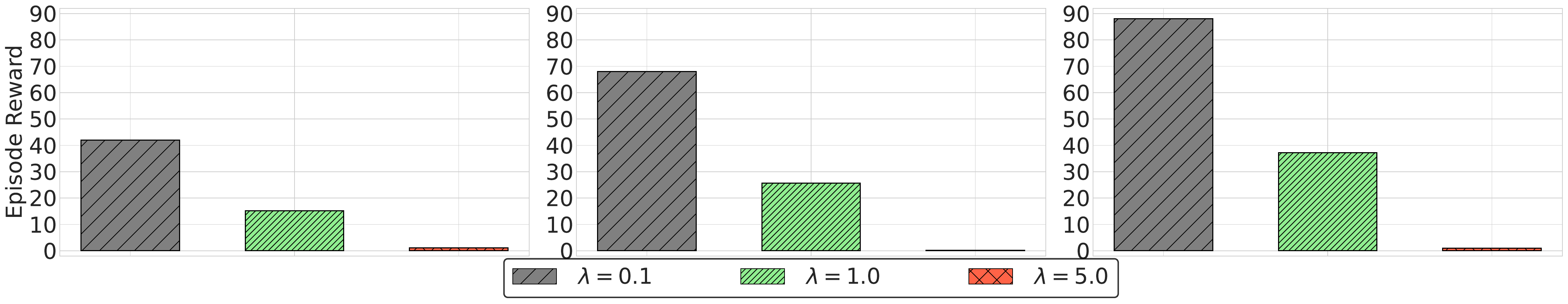}
        \caption{Darkroom. ICPO under different \(\lambda\) scaling the log-probability of \emph{\color{Black}non-preferred} actions with context datasets of {\bf low}, {\bf medium}, and {\bf high} quality.}        
    \end{subfigure}
    \begin{subfigure}{\linewidth}
        \centering
        \includegraphics[width=\linewidth]{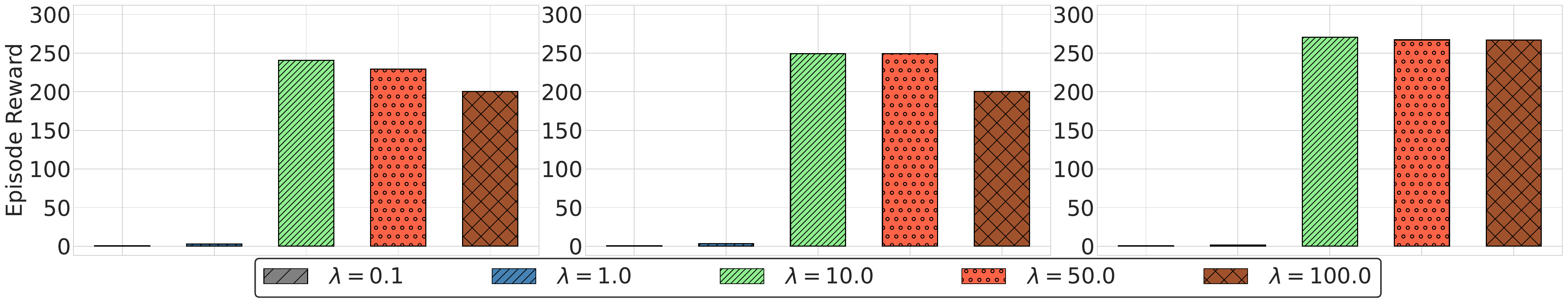}
        \caption{Meta-World. ICPO under different \(\lambda\) that scales the log-probability of \emph{\color{Black}preferred} actions with context datasets of {\bf low}, {\bf medium}, and {\bf high} quality.}
    \end{subfigure}
    \caption{Ablation Studies on the Effect of $\lambda$ for ICPO.}
    \label{fig:lambda_impact}
\end{figure}

This is exactly verified in Figure~\ref{fig:lambda_impact}, as increasing $\lambda$ leads to opposite performance change in DarkRoom and Meta-World. When $\lambda$ is scaling non-preferred actions for DarkRoom, \emph{\color{Black}increasing} $\lambda$ value \emph{\color{Black}decreases} performance. In comparison, when $\lambda$ is scaling preferred actions for Meta-World, \emph{\color{Black}increasing} $\lambda$ value \emph{\color{Black}increases} performance. These results support our insights regarding the effect of $\lambda$. An open question remains: should preferred or non-preferred actions be scaled to achieve better generalization? We leave this investigation to future work.




\section{Use LLMs to label trajectory preference}\label{app:llm}
\begin{figure}[H]
    \centering
    \begin{subfigure}{\linewidth}
        \centering
        \includegraphics[width=0.8\linewidth]{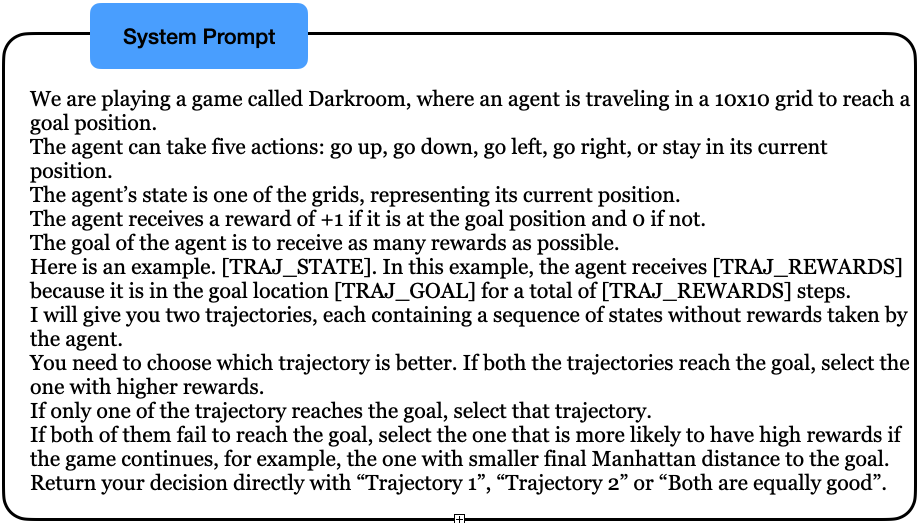}
        \caption{System prompt used for preference labeling with an LLM.}        
    \end{subfigure}
    \begin{subfigure}{\linewidth}
        \centering
        \includegraphics[width=0.8\linewidth]{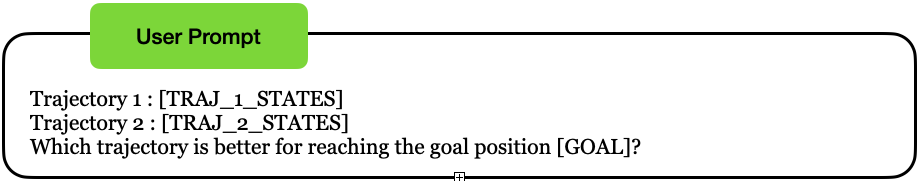}
        \caption{User prompt used for preference labeling with an LLM.}
    \end{subfigure}
    \caption{Prompt input to the LLM for preference labeling.}
    \label{fig:prompt}
\end{figure}

\begin{figure}[H]
    \centering
    \includegraphics[width=0.8\linewidth]{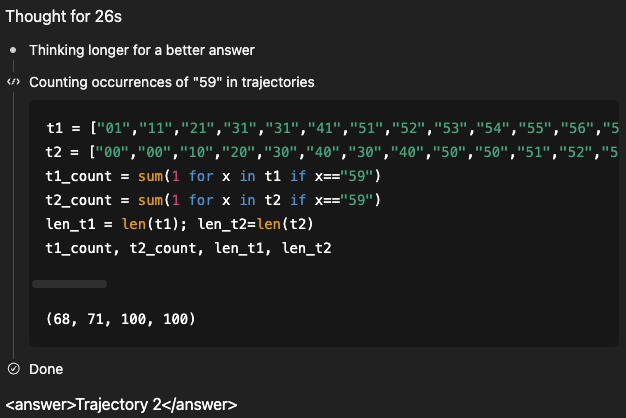}
    \caption{An example response from Chatgpt.}
    \label{fig:chatgpt}
\end{figure}

LLMs are powerful tools for general-purpose tasks. We leverage LLMs to generate preference labels for trajectories in \textbf{Darkroom}, demonstrating that our framework integrates seamlessly with LLMs to scale by reducing the cost of manual preference labeling. Figure~\ref{fig:prompt} illustrates the prompts used for preference labeling. The trajectory states represent the sequence of grid positions visited by the agent, whereas the trajectory goal corresponds to a single state that the LLM leverages for evaluation. We test multiple open-source LLM modes for this purpose, including Qwen2.5-7B-Instruct, Qwen2.5-14B-Instruct, Qwen2.5-32B-Instruct and Qwen2.5-Math-7B-Instruct~\citep{yang2025qwen3}. Surprisingly, these models perform poorly on preference labeling in Darkroom, primarily due to their inability to accurately count the occurrences of goal states. However, this limitation can be readily addressed by leveraging tool-augmented LLMs. We verify that, under the same prompt, \texttt{gpt-4.1-2025-04-14} achieves 100\% accuracy on preference labeling for 100 randomly sampled trajectories pairs, with the labels calculated through the underlying reward function as the true labels. Figure~\ref{fig:chatgpt} illustrates a sample response from ChatGPT, which incorporates a Python snippet to facilitate preference labeling.

\section{Pseudocodes}\label{sec:app-algo}

\begin{algorithm}
\caption{ICRG Pipeline}
\label{alg:icrg}
\begin{algorithmic}[1] 
\STATE {\textbf{Input:} Pretraining dataset $\cD=\{D_i=(\xi^+_i, \xi^-_i)\}$; reward model $R_{\psi}$, policy model $T^R_{\theta}$}
\STATE {\color{red}\fontfamily{cmtt}\selectfont // \; Pretraining for the reward model}
\STATE{Randomly initialize $R_{\psi}$}.
\STATE{Pretrain $R_{\psi}$ with $\cD$ by optimizing the objective function in Equation~\eqref{equ:preference_eqn}.}
\STATE {\color{red}\fontfamily{cmtt}\selectfont // \; Label rewards with the pretrained reward model}
\STATE{Generate rewards for transitions in $\cD$ with $R_{\psi}$ to have a pretraining dataset $\cD^R$ containing trajectories with estimated rewards}.

\STATE {\color{red}\fontfamily{cmtt}\selectfont // \; Pretraining for the policy model with DIT}
\STATE{Randomly initialize the TM policy $T^R_{\theta}$ that requires reward information}.
\STATE{Use the DIT framework (Algorithm~\ref{algo:DIT}) to pretrain $T^R_{\theta}$ with the reward-augmented pretraining dataset $\cD^R$}

\STATE {\color{red}\fontfamily{cmtt}\selectfont // \; Deployment}
\STATE{Upon deployment, receive a pair of preferred and non-preferred trajectories $D^T=(\xi^+, \xi^-)$ and a context dataset $D=\{s_1,a_1,\dots,s_H,a_H\}$ without reward information.}
\STATE{Label all state-action pairs $(s,a)$ in $D$ with $R_\phi(s,a|D^T)$ to have an augmented context dataset $D^R=\{s_1,a_1,\widehat{r}_1,\dots,s_H,a_H,\widehat{r}_H\}$.}
\STATE{Deployment the DIT pretrained TM policy $T^R_\theta(a|s,D^R)$.}
\end{algorithmic}
\end{algorithm}

\section{Computation Resource}\label{sec:app-compute}
All experiments were run on $2$ NVIDIA RTX A6000 GPUs ($48$ GB VRAM) mounted in a $64$-core AMD Threadripper workstation with $256$ GB RAM. Pretraining each transformer model took between $6$ and $12$ hours, depending on the environment and dataset size. Each SAC policy was trained for approximately $3$ hours per task. Across all runs (including pretraining, evaluation, SAC rollouts, and reward estimation), the total compute requirement was approximately $900$ to $1100$ GPU hours.

\end{document}